\documentclass[lettersize,journal]{IEEEtran}
\usepackage{amsmath,amsfonts,amssymb}
\usepackage{algorithmic}
\usepackage{algorithm}
\usepackage{array}
\usepackage[caption=false,font=normalsize,labelfont=sf,textfont=sf]{subfig}
\usepackage{textcomp}
\usepackage{stfloats}
\usepackage{url}
\usepackage{verbatim}
\usepackage{graphicx}
\usepackage{cite}

\usepackage{xcolor}
\usepackage{booktabs} 
\usepackage[normalem]{ulem}
\useunder{\uline}{\ul}{}

\usepackage{booktabs,caption}

\begin{document}

\title{HKAN: Hierarchical Kolmogorov-Arnold Network without Backpropagation}

\author{Grzegorz Dudek,
 Tomasz Rodak
\thanks{G. Dudek is with (i) the Faculty of Electrical Engineering, Czestochowa University of Technology, (ii) the
Faculty of Mathematics and Computer Science, University of Lodz, and (iii) 
the Centre for Data Analysis, Modelling and Computational Sciences (CAMINO), University of Lodz, e-mail: grzegorz.dudek@pcz.pl.}
\thanks{T. Rodak is with the Faculty of Mathematics and Computer Science, University of Lodz, e-mail: tomasz.rodak@wmii.uni.lodz.pl.}}

\markboth{}%
{Shell \MakeLowercase{\textit{et al.}}: A Sample Article Using IEEEtran.cls for IEEE Journals}

\IEEEpubid{0000--0000/00\$00.00~\copyright~2021 IEEE}

\maketitle

\begin{abstract}

This paper introduces the Hierarchical Kolmogorov-Arnold Network (HKAN), a novel neural model that eliminates the need for backpropagation.
While structurally related to the standard Kolmogorov-Arnold Network (KAN), HKAN employs a randomized learning framework and a hierarchical multi-stacking design, where each layer refines the approximations produced by the preceding one through a sequence of convex optimization subproblems.
This non-iterative training strategy ensures high computational efficiency and numerical stability while preserving strong approximation accuracy.
Experimental results on both synthetic and real-world regression tasks show that HKAN achieves accuracy comparable to or exceeding that of standard KANs and Multi-Layer Perceptrons, while reducing training time.
Moreover, HKAN enhances interpretability by incorporating a built-in mechanism for assessing the importance of input variables.
The proposed framework thus bridges theoretical rigor and practical utility, offering a robust, transparent, and computationally efficient alternative to gradient-based neural models.

\end{abstract}


\section{Introduction}
{
\IEEEPARstart{K}{olmogorov-Arnold} Networks (KANs) represent a paradigm shift in neural network (NN) architecture, offering a promising alternative to traditional Multi-Layer Perceptrons (MLPs). Rooted in the Kolmogorov-Arnold representation theorem (KART), which states that any multivariate continuous function $f: [0, 1]^n \to \mathbb{R}$ can be represented as a superposition of continuous functions of a single variable and the binary operation of addition, KANs fundamentally redefine the structure of neural computation.

In this study, we refer specifically to the KAN model introduced by Liu et al. \cite{Liu24}, which sparked the recent surge of research activity in this field. However, the conceptual origins of KANs date back to the 1980s. To the best of our knowledge, an architecture conceptually similar to that of Liu et al., incorporating spline-based basis functions, was first proposed by Igelnik and Parikh in 2003 \cite{Ige03}.
}

KANs introduce a significant architectural innovation: unlike MLPs with fixed activation functions on nodes, KANs employ learnable activation functions on edges. Notably, KANs eliminate linear weights entirely, replacing each weight parameter with a univariate function
{-- specifically, a spline, as in \cite{Liu24}.}
This seemingly simple modification yields significant improvements in both accuracy and interpretability. In terms of accuracy, smaller KAN models consistently achieve comparable or superior performance to larger MLPs in data fitting and partial differential equations solving tasks. Both theoretical analysis and empirical evidence suggest that KANs exhibit faster neural scaling laws than MLPs. Regarding interpretability, KANs offer intuitive visualization and facilitate easy interaction with human users, enhancing their potential as collaborative tools for scientific discovery.

The unique properties of KANs make them valuable collaborators in helping scientists discover mathematical and physical laws, bridging the gap between machine learning and traditional scientific inquiry. As promising alternatives to MLPs, KANs open new avenues for improving contemporary deep learning models, which heavily rely on MLP architectures.

\subsection{Related Work}
{

The origins of KANs can be traced back to the 1980s. Hecht-Nielsen was the first to recognize that KART could be applied in NN computing. He demonstrated that Kolmogorov’s superposition theorem could be interpreted as a four-layer feedforward NN \cite{Hec87}. 
 Kurkova later observed that the fixed number of basis functions in KART could be replaced by a variable number, thereby transforming the task of function representation into one of function approximation \cite{Kur91}. In her subsequent work \cite{Kur92}, she showed how Hecht-Nielsen’s network could be approximated using a traditional NN framework. Sprecher further contributed by deriving numerical algorithms for implementing both the internal and external functions in KART \cite{Spr96,Spr97}.

Igelnik and Parikh later proposed what they termed Kolmogorov’s spline network, effectively an early form of KAN, in which both internal and external functions were represented using cubic spline basis functions \cite{Ige03}. A similar spline-based approach employing cubic B-splines was introduced by Coppejans \cite{Cop04}. Throughout the 1990s and 2000s, various other types of basis functions were explored. For instance, Polar and Poluektov employed piecewise-linear functions whose nodes were adaptively adjusted during training \cite{Pol21}. They demonstrated that the Kolmogorov-Arnold representation is not merely a composition of functions but can also be viewed as a specific case of a tree of discrete Urysohn operators. Moreover, they proposed a fast and computationally stable deep learning algorithm for constructing such Urysohn trees. A practical implementation of a multilayer KAN architecture, along with open-source code, was later provided by the same authors \cite{Pol21a}. In their subsequent work \cite{Pol25}, it was shown that a KAN model trained using the Newton-Kaczmarz method is less sensitive to the choice of the initial guess and learns significantly faster than an MLP while maintaining comparable accuracy.
}

Recent research has highlighted the potential of KANs as efficient and interpretable alternatives to traditional MLPs \cite{Sam24,Pen24,Zhu24}. Unlike MLPs, KANs replace linear weights with learnable activation functions, enabling dynamic pattern learning and improved performance with fewer parameters. Studies have shown that KANs can achieve comparable or even superior accuracy to larger MLPs, faster neural scaling laws, and enhanced interpretability \cite{Sid24}. From a theoretical perspective, Wang et al. \cite{Wan24} demonstrated that the approximation and representation capabilities of KANs are at least equivalent to those of MLPs. Furthermore, KAN’s multi-level learning approach, particularly its grid extension of splines, enhances the modeling of high-frequency components. While MLPs often suffer from catastrophic forgetting, collaborative filtering-based KANs have been proposed to address this issue \cite{Par24}.

Despite these advancements, KANs are not without criticism. Some studies argue that KAN outperforms MLPs primarily in symbolic formula representation but falls short in tasks like computer vision, natural language processing, and audio processing \cite{YuK24}. Tran et al. \cite{Tra24} reported that despite their theoretical advantages, KANs do not consistently outperform MLPs in practical classification tasks. Additionally, their hardware implementations tend to be less efficient, with higher resource usage and latency. Sensitivity to noise {can be} another limitation; even minimal noise in the data can significantly degrade performance \cite{She24}.

To enhance the interpretability, interactivity, and versatility of KAN, Liu et al. \cite{Liu24b} introduced MultKAN, which incorporates multiplication operations. By integrating multiplication nodes, MultKAN explicitly represents multiplicative structures, allowing for a more transparent mapping of physical laws and improved modeling of complex relationships.

KANs have also been integrated with other architectures to address diverse challenges. In computer vision, \cite{Bod24} combined KANs with convolutional layers, demonstrating that KAN convolutions maintain similar accuracy while using half the parameters. Residual KANs, introduced in \cite{YuR24}, effectively capture long-range, nonlinear dependencies within CNNs by incorporating KAN as a residual component. 

Genet and Inzirillo \cite{Gen24b} proposed integrating KANs with transformers to simplify complex dependencies in time series while enhancing interpretability. 
Temporal KANs, combining KAN with LSTMs, were introduced in \cite{Gen24a} for multi-step time series forecasting. These networks integrate memory management through recurrent KAN layers. Subsequent work by the same authors refined this approach by incorporating transformers and learnable path signatures to capture geometric features \cite{Inz24}. 

Yang and Wang \cite{Yan24} introduced the Kolmogorov-Arnold transformer, replacing MLP layers with KAN layers to improve model expressiveness and performance.
In graph NNs networks, \cite{Zha24} replaced MLPs with KANs for feature extraction, resulting in the GraphKAN architecture. Li et al. \cite{LiG24} tailored a KAN-GNN model for molecular representation learning, emphasizing KAN’s flexibility in diverse domains.

KANs have also been employed in evolutionary algorithms as surrogate models for regression and classification tasks \cite{Hao24}, helping to reduce the number of expensive function evaluations during optimization. Additionally, \cite{Che24} introduced probabilistic KANs by incorporating Gaussian process neurons, enabling robust nonlinear modeling with uncertainty estimation.

The flexibility of KANs has led to explorations with various activation functions beyond B-splines, including {piecewise-linear functions \cite{Pol21}}, wavelets \cite{Boz24}, radial basis functions \cite{Abu24}, Fourier series \cite{Xu24}, Jacobi basis functions \cite{Agh24}, rational functions \cite{AghR24}, and ReLU \cite{Qui24}. 
A comprehensive comparison of activation functions used in KAN architectures is available in \cite{Ta24}.


Numerous enhancements have been proposed for KANs, such as dropout-based regularization \cite{Alt24}, adaptive grid updates \cite{Rig24}, federated learning \cite{Zey24}, and reinforcement learning \cite{Kic24}. These advancements, combined with KAN’s interpretability and flexibility, have enabled its application across a wide range of fields, including tabular data \cite{Poe24}, computer vision \cite{Li24, Bod24}, graphs \cite{Zha24}, time series \cite{Vac24, Xu24b, Zho24, Gen24a}, recommender systems \cite{Par24}, neuroscience \cite{YanE24}, quantum science \cite{Kun24}, biology \cite{Pra24}, and survival analysis \cite{Kno24}.

\subsection{Motivation and Contributions}

{KAN models are traditionally trained using the backpropagation algorithm.
In this work, we use the term backpropagation in its broader, commonly accepted sense -- that is, as the core learning mechanism typically employed to train NNs by minimizing a loss function through gradient-based optimization. In this context, it encompasses both the computation of gradients and their use by an optimization algorithm (e.g., stochastic gradient descent) to iteratively update model parameters across layers.}

{However, gradient-based learning processes are prone to challenges such as local minima, flat regions, and saddle points in the loss landscape. In addition, gradient computation can be computationally demanding, particularly for deep or wide architectures, complex target functions, and large-scale datasets.
These methods may also suffer from the vanishing or exploding gradient problem, which can significantly hinder convergence and stability in deep NNs. 
}

In this study, we propose a randomized learning approach for training KANs as an alternative to backpropagation. Unlike gradient-based methods, which lead to non-convex optimization problems, the randomized approach transforms the problem into a convex one \cite{Pri15}. This is achieved by fixing the parameters of the activation functions, which are selected either randomly or based on the data, and remain unchanged during training. The only adaptation occurs in the linear functions that aggregate the outputs of the basis functions and activation functions. Since the optimization problem becomes linear, the model’s weights can be efficiently learned using a standard least-squares method. This significantly simplifies the training process and accelerates computation compared to gradient-based approaches. Numerous studies in the literature have demonstrated the high performance of randomized neural models compared to fully trainable ones \cite{Pao94, Nee20, Zha16, Sca17, Mal23, Dud19, Dud20, Dud21}. 

Our approach begins with utilizing fixed parameters for basis functions, determined either by data or randomly. These basis functions are then combined in multiple blocks using linear regression. 
The resulting block functions (activation functions) are subsequently combined through linear regression, and this iterative process is repeated across subsequent layers to form higher-level representations {(such layer-wise learning -- progressively improving higher-level abstractions of the input and enhancing generalization -- has been employed before, albeit in a fundamentally different manner, for instance, in the greedy layer-wise unsupervised training strategies used in Deep Belief Networks \cite{Hin06,Ben06}).} 
Combining diverse blocks corresponds to ensembling, while performing it layer by layer constitutes a multi-stacking approach. 
This hierarchical modeling of the target function progressively enhances accuracy at each level, eliminating the need for backpropagation.

Our study makes tree significant contributions to the field of NNs, specifically in the domain of KANs:
\begin{enumerate}
 \item{Novel Training Method for KAN}: 
 We introduce an innovative approach to training KANs that eliminates the need for backpropagation. The parameters of basis functions are fixed, determined either randomly or based on data. The model is trained hierarchically using the standard least-squares method. This approach results in a more efficient and robust training process for KANs, offering improvements in both computational efficiency and model accuracy.
 \item{Multi-Stacking Approach for Prediction}:
 Our hierarchical KAN implements hierarchically multi-stacking approach to built predictions. In each layer, meta-learners combine predictions performed by weak learners (univariate models). Subsequent layers, fed by predictions from previous layers, successively refine the results, enhancing overall accuracy layer by layer.
 \item{Empirical Results for Regression Problems}: 
 We provide comprehensive empirical evidence demonstrating that our HKAN outperforms standard KAN {and MLP} in a range of regression problems.
\end{enumerate}

The remainder of this paper is organized as follows:  
Section II provides an overview of the Kolmogorov-Arnold representation theorem and standard KANs, establishing the foundation for our research.  
Section III introduces the proposed HKAN model, detailing its architecture, components, features, and learning process.  
Section IV presents a comparison between HKAN and KAN, while Section V examines HKAN through the lens of multi-stacking models.  
The experimental framework used to evaluate the proposed model is described in Section VI.  
Finally, Section VII concludes the paper.  

\section{Preliminary}

\subsection{Kolmogorov-Arnold Representation Theorem}

KART, also known as the superposition theorem, stands as a cornerstone in the theory of function approximation. This profound result asserts that any continuous function of several variables can be represented as a composition of continuous functions of one variable and addition.

For any continuous function $f: [0,1]^n \rightarrow \mathbb{R}$, there exist continuous functions $\phi_{q,p}: [0,1] \rightarrow \mathbb{R}$ and $\Phi_q: \mathbb{R} \rightarrow \mathbb{R}$ such that:

\begin{equation}
\label{eq1}
f(x_1, \ldots, x_n) = \sum_{q=1}^{2n+1} \Phi_q \left( \sum_{p=1}^n \phi_{q,p}(x_p) \right)
\end{equation}

The theorem carries significant implications for function approximation and theoretical computer science. It suggests a universal approximation capability, implying that any multivariate continuous function can be approximated by a network of simple, single-variable functions. Notably, the outer functions $\Phi_q$ are independent of the function $f$ being approximated, serving as universal building blocks. This property effectively reduces the problem of approximating $n$-dimensional functions to that of approximating one-dimensional functions.

However, the theorem's practical application faces certain limitations. The inner functions $\phi_{q,p}$ can be highly non-smooth, even when $f$ is smooth, potentially complicating computational implementation. Moreover, while theoretically powerful, the representation may not be efficiently computable in practice.

KART stands as a bridge between pure mathematics and applied computational science, highlighting the potential for representing complex functions through simpler components while also illustrating the challenges in translating theoretical results into practical applications.

\subsection{Kolmogorov-Arnold Networks (KANs)}

Paper \cite{Liu24} extends and modifies KART to create KANs in several key ways. While the original theorem uses a 2-layer network with a specific width in the hidden layer, KANs generalize this to allow arbitrary widths and depths, stacking multiple "KAN layers".  

A KAN layer is defined as a matrix of activation functions $\phi_{q,p}$, where $q$ is not restricted to the theoretical limit of $2n+1$: 

\begin{equation}
\label{eq2}
\phi(\mathbf{x})= \begin{bmatrix} \phi_{1,1}(x_1) & \ldots & \phi_{1,n_{in}} (x_{n_{in}})\\ \vdots & \ddots & \vdots \\ \phi_{n_{out},1}(x_1) & \ldots & \phi_{n_{out},n_{in}}(x_{n_{in}}) \end{bmatrix}
\end{equation}
where $x_p$ denotes the $p$-th input to the layer, $n_{in}$ denotes the number of inputs, and $n_{out}$ denotes the number of outputs (not restricted to $2n+1$). 

Deeper KANs are created by composing multiple KAN layers. Unlike the original theorem which allows non-smooth or even fractal functions, KANs assume smooth activation functions to facilitate learning. The authors propose activation functions parameterized as B-splines with trainable coefficients combined with the sigmoid linear unit (SiLU). Note the substantial difference between KANs and MLPs: instead of fixed multidimensional activation functions on nodes as in MLPs, KANs use learnable one-dimensional activation functions on edges.

The paper introduces a grid extension technique, allowing KANs to be made more accurate by refining the spline grids of the activation functions. This enables increasing model capacity without retraining from scratch.

The authors also introduce sparsification and pruning techniques to simplify KANs and discover minimal architectures that match the data structure. New theoretical guarantees are provided for KANs with finite grid sizes, suggesting that they can beat the curse of dimensionality for functions with compositional structure.

Furthermore, the paper provides tools for users to visualize and modify KANs, making them more interpretable and interactive. This approach takes the core idea of representing multivariate functions using univariate functions and addition, and extends it into a flexible, trainable NN architecture with theoretical guarantees and practical advantages over standard MLPs. In essence, \cite{Liu24}  modernizes KART for use in contemporary machine learning, offering a new perspective on function approximation and NN design.


\section{Hierarchical KAN}

Table \ref{tab1} provides a summary of the main symbols used throughout this study for clarity and reference. The implementation of the proposed model is available in our GitHub repository \cite{Rod25}.

\begin{table}[!t]
 \setlength{\tabcolsep}{5pt}
 \caption{List of the Main Symbols.}
 \label{tab1}
 \centering
 \begin{tabular}{cl}
\hline
  Symbol& Meaning \\ 
\hline
$\text{BaF}$ & basis function\\
$\text{BlF}$ & block function (activation function)\\
$n$ & number of inputs	\\
$N$ & number of training samples	\\
$\mathbf{x} \in [0, 1]^n$ & input pattern		\\
$\mathbf{z}^{(l)} \in \mathbb{R}^{n^{(l)}}$ & output vector of layer $l$ and input vector to layer $l+1$ 		\\
$y \in [0, 1]$ & target		\\
$\hat{y}\in \mathbb{R}$ & prediction \\
$l, L$ & layer index and the total number of layers, respectively\\
$n^{(l)}$ & width of the $l$-th layer (number of nodes)\\
$m^{(l)}$ & 
number of BaFs in a block of layer $l$\\ 
$p$ & input index to layer $l$, $p=1,...,n^{(l-1)}$\\
$q$ & output index of layer $l$, $q=1,...,n^{(l)}$\\
$r$ & BaF index in a block in layer $l$, $r=1,...,m^{(l)}$\\
$g^{(l)}_{q,p,r}$ & BaF in layer $l$\\  
$\phi^{(l)}_{q,p}$ & BlF in layer $l$, i.e. linear combination of  $g^{(l)}_{q,p,r}$\\
$h^{(l)}_{q}$ & $q$-th output of layer $l$, i.e. linear combination of $\phi^{(l)}_{q,p}$  \\
$c^{(l)}_{q,p,r}$ & weight of BaF $g^{(l)}_{q,p,r}$\\
$w^{(l)}_{q,p}$ & weight of BlF $\phi^{(l)}_{q,p}$\\
$\mu^{(l)}_{q,p,r}$ & location parameter of BaF $g^{(l)}_{q,p,r}$\\
$\sigma^{(l)}$ & smoothing parameter in layer $l$\\
$\lambda_{\phi}^{(l)}, \lambda_h^{(l)}$ & regularization parameters for functions $\phi$ and $h$ in layer $l$\\
\hline
\end{tabular}
\end{table}

{
\subsection{Model} \label{Arch}

HKAN is an advanced NN architecture inspired by KART. While it shares similarities with the KAN architecture, HKAN introduces unique components and a distinct training process. The architecture of HKAN is shown in Fig. \ref{fig1}. 

\begin{figure}[!t]
\centering
\includegraphics[width=0.5\textwidth]{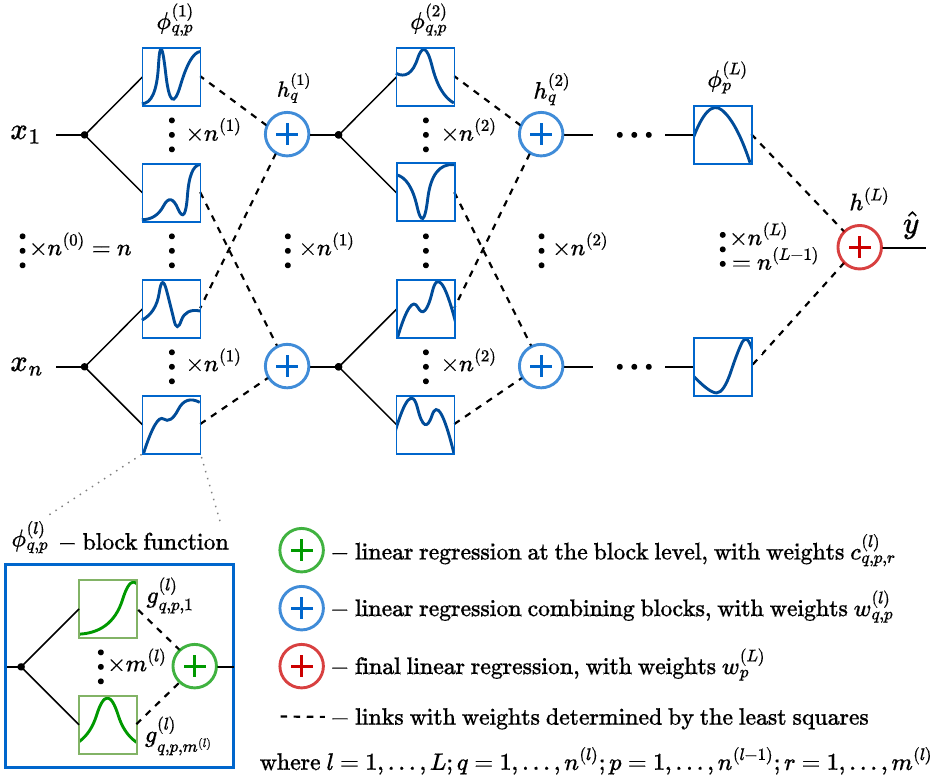}
\caption{HKAN architecture.}
\label{fig1}
\end{figure}

Let $\mathbf{z}^{(l-1)}$ denote the input vector to layer $l$.
The first layer of HKAN receives the input vector $\mathbf{z}^{(0)} = \mathbf{x} = [x_1, \ldots, x_n]^\top \in \mathbb{R}^n$.
Each component of the input vector is processed by a corresponding group of blocks. Within each group, there are $n^{(l)}$ blocks, and each block applies a nonlinear projection of its input using $m^{(l)}$ basis functions (BaFs).
Consider two common basis functions: the Gaussian function

\begin{equation}
\label{eq3}
g(z) = \exp \left(-\left( \sigma(z-\mu)\right)^2\right)
\end{equation}
and the sigmoid function

\begin{equation}
\label{eq4}
g(z) = \frac{1}{1+\exp \left(-\sigma(z-\mu)\right)}
\end{equation}
where $\mu$ is the location parameter, and $\sigma$ is the smoothing parameter corresponding to the slope or bandwidth of the BaF. 

The BaFs within a block, denoted $g^{(l)}_{q,p,r}: \mathbb{R} \rightarrow \mathbb{R}$, are combined linearly :

\begin{equation}
\label{eq5}
\phi^{(l)}_{q,p}(z^{(l-1)}_p) = \sum_{r=1}^{m^{(l)}} c^{(l)}_{q,p,r}g^{(l)}_{q,p,r}(z^{(l-1)}_p)
\end{equation}

The resulting function $\phi^{(l)}_{q,p}: \mathbb{R}^{m^{(l)}} \rightarrow \mathbb{R}$ is referred to as a block function (BlF), corresponding to the inner function in \eqref{eq1}.
Fig.~\ref{fig2} shows the composition of a block function using BaFs.

\begin{figure}[!t]
\centering
\includegraphics[width=0.244\textwidth]{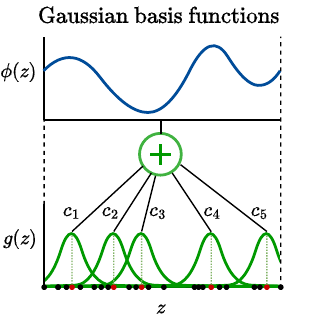}
\includegraphics[width=0.238\textwidth]{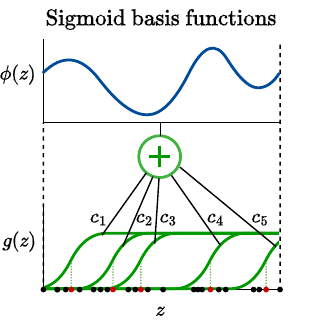}
\caption{Illustration of block function composition using  BaFs.}
\label{fig2}
\end{figure}

Subsequently, the $n^{(l-1)} n^{(l)}$ BlFs are linearly combined by $n^{(l)}$ $h$-functions, corresponding to the outer functions defined in \eqref{eq1}.
Each $q$-th $h$-function takes as input the $q$-th BlF from every group:

\begin{equation}
\label{eq9}
h^{(l)}_{q}(\mathbf{z}^{(l-1)}) = \sum_{p=1}^{n^{(l-1)}} w^{(l)}_{q,p}\phi^{(l)}_{q,p}(z^{(l-1)}_p)
\end{equation}

The output of layer $l$ is given by $\mathbf{z}^{(l)}=\hat{\mathbf{y}}^{(l)}=[\hat{y}^{(l)}_1, ..., \hat{y}^{(l)}_{n^{(l)}}]^\top \in \mathbb{R}^{n^{(l)}}$, where each component $\hat{y}^{(l)}_q = h^{(l)}_q(\mathbf{z}^{(l-1)})$. This output is then fed to the next layer and processed using \eqref{eq5} and \eqref{eq9}.

Following the KAN design introduced in \cite{Liu24}, the structure of the top layer ($L$) in HKAN slightly differs from that of the preceding layers, as it contains only one BaF per input. The final output of HKAN is then produced as a linear combination of the $n^{(L)} = n^{(L-1)}$ BlFs:

\begin{equation}
\label{eq9.5}
h^{(L)}(\mathbf{z}^{(L-1)}) = \sum_{p=1}^{n^{(L)}} w^{(L)}_{p}\phi^{(L)}_{p}(z^{(L-1)}_p)
\end{equation}

\subsection{Learning}

HKAN learning is hierarchical and proceeds layer by layer. Below, we discuss the parameter selection/estimation process for BaFs, BlFs, and $h$-functions. The parameters of the BaFs are determined in a randomized manner (without learning), whereas for the BlFs and $h$-functions, parameter estimation is performed using linear regression.

\subsubsection{\textbf{BaFs}}

The configuration of BaFs plays a crucial role in the network's performance. The number of BaFs, together with the smoothing parameter, serve as hyperparameters that define the block's flexibility and balance the trade-off between variance and bias in the output.
The locations of the BaFs (denoted as $\mu^{(l)}_{q,p,r}$ for the $r$-th function in the block $\phi^{(l)}_{q,p}$) define the position of the maximum for Gaussian functions or the inflection point for sigmoid functions when these two basis functions are considered.
To distribute BaFs in a block across the input interval (typically a bounded region of $[0, 1]$), these locations can be selected in two ways \cite{Dud19,Dud20}: 
\begin{itemize} 
\item Random uniform distribution: The locations are drawn from a uniform distribution, $\mu^{(l)}_{q,p,r} \sim U(0,1)$, ensuring random spread across the input range. 
\item Data-driven distribution (support point method): In this approach, locations are assigned to randomly selected training points (in the first layer) or their projections (in subsequent layers), called support points: $\mu^{(l)}_{q,p,r} = z^{(l-1)}_{\xi,p}$, where $\xi \sim U\{1, .., N\}$. 
\end{itemize}

The support point method aligns the BaFs with the data distribution, thereby avoiding empty regions in the input space. In Fig. \ref{fig2}, the red markers represent the support points that determine the placement of the BaFs.

\subsubsection{\textbf{BlFs}}

The weights of the BaFs in \eqref{eq5}, denoted as $c^{(l)}_{q,p,r}$, are determined using the least-squares method by minimizing the sum of squared residuals:

\begin{equation}
\label{eq6}
L = \sum_{i=1}^{N} (y_i-\hat{y}_i)^2
\end{equation}
where $\hat{y}_i$ is a prediction performed by the BlF: $\hat{y}_i=\phi^{(l)}_{q,p}(z^{(l-1)}_{i,p})$.

The weights of $\phi^{(l)}_{q,p}$  can be calculated analitically as

\begin{equation}
\label{eq7}
\mathbf{c}^{(l)}_{q,p} = \mathbf{G}^{(l)+}_{q,p}\mathbf{y}
\end{equation}
where $\mathbf{c}^{(l)}_{q,p}=[c^{(l)}_{q,p,1}, ..., c^{(l)}_{q,p,m^{(l)}}]^\top$, $\mathbf{y}=[y_1, ..., y_N]^\top$ and $\mathbf{G}^{(l)+}_{q,p}$ is the Moore-Penrose generalized inverse of the BaF response matrix to the $N$ training data points or their projections:

\begin{equation}
\label{eq8}
\mathbf{G}^{(l)}_{q,p}= \begin{bmatrix} g_{q,p,1}(z^{(l-1)}_{p,1}) & \ldots & g_{q,p,m^{(l)}}(z^{(l-1)}_{p,1})\\ \vdots & \ddots & \vdots \\ g_{q,p,1}(z^{(l-1)}_{p,N})  & \ldots & g_{q,p,m^{(l)}}(z^{(l-1)}_{p,N}) \end{bmatrix}
\end{equation}

\subsubsection{\textbf{h-functions}}

The weights of combination \eqref{eq9} are calculated as:

\begin{equation}
\label{eq10}
\mathbf{w}^{(l)}_{q} = \mathbf{\Phi}^{(l)+}_{q}\mathbf{y}
\end{equation}
where $\mathbf{w}^{(l)}_{q}=[w^{(l)}_{q,1}, ..., w^{(l)}_{q,n^{(l)}}]^\top$, and $\mathbf{\Phi}^{(l)+}_{q}$ is the Moore-Penrose generalized inverse of the BlF response matrix to the projections of $N$ training data points:

\begin{equation}
\label{eq11}
\mathbf{\Phi}^{(l)}_{q}= \begin{bmatrix} \phi_{q,1}(z^{(l-1)}_{1,1}) & \ldots & \phi_{q,n^{(l-1)}}(z^{(l-1)}_{{n^{(l-1)},1}})\\ \vdots & \ddots & \vdots \\ \phi_{q,n^{(l)}}(z^{(l-1)}_{1,N})  & \ldots & \phi_{q,n^{(l)}}(z^{(l-1)}_{{n^{(l-1)},N}}) \end{bmatrix}
\end{equation}

Weights \eqref{eq10} minimize loss function \eqref{eq6}, where $\hat{y}_i$ is a prediction performed by the $h$-function: $\hat{y}_i=h^{(l)}_{q}(\mathbf{z}^{(l-1)}_i)$.

In HKAN, we employ standard linear regression, which is applied multiple times both at the block level (BlFs) and for combining multiple blocks ($h$-functions). However, to mitigate overfitting, we can alternatively use regularized least squares (ridge regression). In the experimental part of this work, we adopt this variant to calculate the weights of BlFs, $\phi^{(l)}_{q,p}$. The weights in this case are computed using the following closed-form solution:

\begin{equation}
\label{eq7a}
\mathbf{c}^{(l)}_{q,p} = (\mathbf{G}^{(l)\top}_{q,p}\mathbf{G}^{(l)}_{q,p}+\lambda_{\phi}^{(l)}\mathbf{I})^{-1}\mathbf{G}^{(l)\top}_{q,p}\mathbf{y}
\end{equation}
where $\mathbf{I}$ is an identity matrix and $\lambda_{\phi}^{(l)} \geq 0$ is a regularization parameter.

The optimization problem in HKAN is decomposed into multiple linear regression subproblems. Each subproblem minimizes objective function \eqref{eq6} using the least squares method. Since these optimization subproblems are convex, 
the least squares approach guarantees optimal weights (within the context of the randomly selected BaFs).

The learning process in HKAN is hierarchical. First, the target function is simultaneously approximated by the blocks of the first layer, with each block modeling the target based on a single input variable. Due to the nonlinear nature of the BaFs, this step introduces nonlinearity into the modeling process. Then, based on these approximations, multiple $h$-functions are fitted to the target function. Each $h$-function linearly combines single-variable BlFs, producing a multivariable mapping.

Subsequent layers transform their inputs in a similar manner, involving three key steps: (1) nonlinear projections of individual inputs by BaFs, (2) linear combinations of BaFs within each block, and (3) linear combinations of BlFs. Each layer refines the predictions generated by the preceding layers, aiming to improve the overall approximation.

It is important to note that BaFs are not learned; their parameters, namely location ($\mu$) and smoothing ($\sigma$), remain fixed. Randomness in $\mu$ introduces diversity among BlFs.  
This diversity is advantageous for ensembling, which is carried out by the $h$-functions. The benefits of this approach are discussed in greater detail in Section \ref{MSM}.

Table~\ref{tabL} summarizes the HKAN parameters and the stages of the hierarchical learning process.

\begin{table*}[]
\caption{{Overview of the HKAN parameters and the sequential stages of its hierarchical learning process.}}
\label{tabL}
\setlength{\tabcolsep}{7pt}
\centering
\begin{tabular}{l|ccc}

\multicolumn{4}{c}{\textbf{Layer 1}}  \\\hline 
HKAN component & \multicolumn{1}{c|}{BaFs}  & \multicolumn{1}{c|}{BlFs} & $h$-functions \\ 
Parameters & \multicolumn{1}{c|}{$\mu^{(1)}_{q,p,r}$} & \multicolumn{1}{c|}{$c^{(1)}_{q,p,r}$}  & $w^{(1)}_{q,p}$ \\ 
Number of parameters& \multicolumn{1}{c|}{$n \cdot n^{(1)} \cdot m^{(1)}$}   & \multicolumn{1}{c|}{$n \cdot n^{(1)} \cdot m^{(1)}$}  & $n \cdot n^{(1)}$  \\ 

Learning method  & \multicolumn{1}{c|}{None; fixed parameters} & \multicolumn{1}{c|}{Least squares with target $y$} & \multicolumn{1}{c}{Least squares with target $y$} \\   
&\multicolumn{1}{c|}{(random/data-driven) }&\multicolumn{1}{c|}{} & \multicolumn{1}{c}{} \\

Number of least-squares runs & \multicolumn{1}{c|}{-}  & \multicolumn{1}{c|}{$n \cdot n^{(1)}$}  & $n^{(1)}$ \\ 

Output& \multicolumn{1}{c|}{$g^{(1)}_{q,p,r}$ -- nonlinear randomized} & \multicolumn{1}{c|}{$\phi^{(1)}_{q,p}$ -- linear combinations} & $h^{(1)}_q$ -- linear combinations \\  
& \multicolumn{1}{c|}{projections of inputs $x_p$} & \multicolumn{1}{c|}{of BaFs in each block} & of selected BlFs \\
\hline
\multicolumn{4}{l}{}  \\
\multicolumn{4}{c}{\textbf{Layer 2}}  \\\hline 
HKAN component & \multicolumn{1}{c|}{BaFs}  & \multicolumn{1}{c|}{BlFs} & $h$-functions \\  
Parameters & \multicolumn{1}{c|}{$\mu^{(2)}_{q,p,r}$} & \multicolumn{1}{c|}{$c^{(2)}_{q,p,r}$}  & $w^{(2)}_{q,p}$ \\ 
Number of parameters& \multicolumn{1}{c|}{$n^{(1)} \cdot n^{(2)} \cdot m^{(2)}$}  & \multicolumn{1}{c|}{$n^{(1)} \cdot n^{(2)} \cdot m^{(2)}$} & $n^{(1)} \cdot n^{(2)}$ \\ 
Learning method  & \multicolumn{1}{c|}{None; fixed parameters} & \multicolumn{1}{c|}{Least squares with target $y$} & \multicolumn{1}{c}{Least squares with target $y$} \\   
&\multicolumn{1}{c|}{(random/data-driven) }&\multicolumn{1}{c|}{} & \multicolumn{1}{c}{} \\

Number of least-squares runs & \multicolumn{1}{c|}{-}  & \multicolumn{1}{c|}{$n^{(1)} \cdot n^{(2)}$} & $n^{(2)}$ \\

Output& \multicolumn{1}{c|}{$g^{(2)}_{q,p,r}$ -- nonlinear randomized} & \multicolumn{1}{c|}{$\phi^{(2)}_{q,p}$ -- linear combinations} & $h^{(2)}_q$ -- linear combinations \\  
& \multicolumn{1}{c|}{projections of inputs $z^{(1)}_p$} & \multicolumn{1}{c|}{of BaFs in each block} & of selected BlFs \\
\hline
\multicolumn{4}{l}{}  \\
\multicolumn{4}{c}{\textbf{Layer 3 (top)}}  \\\hline 
HKAN component & \multicolumn{1}{c|}{BaFs}  & \multicolumn{1}{c|}{BlFs} & $h$-functions \\  
Parameters & \multicolumn{1}{c|}{$\mu^{(3)}_{q,p,r}$} & \multicolumn{1}{c|}{$c^{(3)}_{q,p,r}$}  & $w^{(3)}_{p}$ \\ 
Number of parameters& \multicolumn{1}{c|}{$n^{(3)} \cdot m^{(3)}$}  & \multicolumn{1}{c|}{$n^{(3)} \cdot m^{(3)}$} & $n^{(3)}$ \\ 
Learning method  & \multicolumn{1}{c|}{None; fixed parameters} & \multicolumn{1}{c|}{Least squares with target $y$} & \multicolumn{1}{c}{Least squares with target $y$} \\   
&\multicolumn{1}{c|}{(random/data-driven/even distrib.) }&\multicolumn{1}{c|}{} & \multicolumn{1}{c}{} \\
Number of least-squares runs & \multicolumn{1}{c|}{-}  & \multicolumn{1}{c|}{$n^{(2)} \cdot n^{(3)}$} & $1$  \\ 

Output& \multicolumn{1}{c|}{$g^{(3)}_{p,r}$ -- nonlinear randomized} & \multicolumn{1}{c|}{$\phi^{(3)}_{p}$ -- linear combinations} & $h^{(3)}$ -- linear combinations \\  
& \multicolumn{1}{c|}{projections of inputs $z^{(2)}_p$} & \multicolumn{1}{c|}{of BaFs in each block} & of BlFs \\
 \hline
\end{tabular}
\end{table*}

}



\subsection{Hyperparameters}

The HKAN hyperparameters are as follows:

\begin{itemize}
 \item $L$ -- total number of layers,  
 \item $n^{(l)}$ -- width of the $l$-th layer (number of nodes), 
 \item $m^{(l)}$ -- number of BaFs in blocks of layer $l$,
 \item BaF type,
 \item Way to generate location parameters of BaFs ($\mu$),
 \item $\sigma^{(l)}$ -- smoothing parameter in layer $l$,
 \item $\lambda_{\phi}^{(l)}, \lambda_h^{(l)}$ -- regularization parameters for functions $\phi$ and $h$ in layer $l$ (optional).
\end{itemize}

Intuitively, a more complex target function needs a deeper and wider network to model it with higher accuracy. The modeling burden can be shifted at the block level. In such a case many BaFs are needed (with $\sigma$-parameter adjusted to the approximation problem complexity) and less $h$-functions. Opposite situation is also possible: a small number of BaFs roughly approximates the target function, and the effort of a more accurate approximation falls on a large number of $h$-functions. 

It is important to note that the regression problem solved at each level of HKAN processing can vary. At the initial level, the target function is approximated based directly on input patterns $\mathbf{x}$, whereas at subsequent levels, it is approximated based on the predictions generated by the previous level. Consequently, the complexity of the problems addressed at different levels may differ, necessitating distinct values for layer-specific hyperparameters $n^{(l)}$, $m^{(l)}$, $\sigma^{(l)}$, and optionally $\lambda_{\phi}^{(l)}$ and $\lambda_h^{(l)}$.

HKANs with more layers, more nodes, more BaFs and with smaller values of the parameters $\sigma$ and $\lambda$ tend to fit the target function more accurately. 
However, such configurations are more susceptible to overfitting. Therefore, these parameters must be tuned to strike an optimal balance between the model's bias and variance.

The strategy for generating $\mu$ values, which determine the positions of BaFs, depends on the anticipated data distribution. When the distribution of new, unseen data points is expected to closely mirror that of the training dataset, positioning BaFs at the training points is often an effective approach. This method ensures that the network's approximation capability is concentrated in regions where data is most likely to occur. Conversely, if the distribution of new data is expected to differ from the training set, or if the goal is to create a more generalized model, a random distribution of BaFs may be preferable. This approach allows for broader coverage of the input space, including regions that may be sparsely represented or entirely absent in the training data. 

Determining the optimal shape of BaFs a priori is challenging, as it often depends on the specific characteristics of the target function. In HKAN, different types of BaFs can be mixed flexibly. They may vary across layers, between blocks within the same layer, or even within individual blocks. In such cases, the smoothing parameters should be customized for each type of BaF to ensure optimal performance and adaptability.





\subsection{Complexity} \label{comp}

In layer $l$, each block performs linear regression on $m^{(l)}$ BaFs with a complexity of $O(Nm^{(l)2} + m^{(l)3})$.
For layers $l = 1, ..., L-1$, each containing $n^{(l-1)}n^{(l)}$ blocks, the total complexity per layer is $O(n^{(l-1)}n^{(l)}(Nm^{(l)2} + m^{(l)3}))$. The top layer contains $n^{(L)}$ blocks, resulting in a complexity of $O(n^{(L)}(Nm^{(L)2} + m^{(L)3}))$.

Each function $h^{(l)}_q$ linearly combines $n^{(l-1)}$ blocks, yielding a complexity of $O(Nn^{(l-1)2} + n^{(l-1)3})$. For layers $l = 1, ..., L-1$, with $n^{(l)}$ such functions per layer, the total complexity becomes $O(n^{(l)}(Nn^{(l-1)2} + n^{(l-1)3}))$. The final layer produces a single output, resulting in a complexity of $O(Nn^{(L)2} + n^{(L)3})$.

\section{HKAN vs Standard KAN}

This section outlines the key differences between KAN as defined in \cite{Liu24} and our proposed HKAN.

\subsection{Basis Functions}
KAN {introduced in \cite{Liu24}} employs B-splines of order 3, which are bell-shaped and computed recursively using the Cox-de Boor formula. The properties of these BaFs, including their number, location, width, and support, are determined by knots. In KANs, these knots are positioned at equidistant intervals, resulting in an even distribution of the basis functions across the input space. The number of knots, which directly influences the spline's flexibility and the model's capacity to capture complex patterns, is a crucial hyperparameter in the KAN architecture. 

In contrast, HKANs offer greater flexibility in the selection of BaF types. While in Section \ref{Arch} we introduced Gaussian and sigmoid functions, the HKAN framework is not limited to these and can accommodate various other functional forms (see experimental part of this work, Section \ref{ExSt}). Unlike KANs, the distribution of BaFs in HKANs can be either data-driven or random. In the HKAN framework, the smoothing parameter and the number of BaFs serve as key hyperparameters. The adaptability in both function type and distribution allows HKANs to potentially capture a wider range of functional relationships within the data.

When evaluating computational efficiency, it should be noted that KAN uses B-splines, whose computation involves recursive processes, making it computationally intensive. In contrast, HKAN does not require a recursive process to create BaFs, potentially reducing computational complexity.

\subsection{Block Functions (Activation Functions)}
In KAN, what we refer to as a BlF is termed an activation function. This activation function is a composite structure, consisting of two main components: a weighted sum of a spline (which itself is a linear combination of BaFs with trainable weights $c$) and a SiLU. The incorporation of SiLU was likely designed to enhance the network's training dynamics. In the KAN architecture, the BlFs are aggregated without additional weighting.

HKAN employ a distinct methodology. In this framework, BlFs are also constructed by combining BaFs with weights $c$, similar to KAN. However, unlike in KAN, these weights are not optimized via gradient descent. Instead, they are computed using the least-squares method, with the goal of fitting each BlF to the target function in a one-dimensional space. This method provides a more direct, analytical determination of the weights. The HKAN then linearly combines these BlFs, once again using weights determined through the least-squares method, to approximate the target function in multi-dimensional space.

\subsection{Explainability and Function Representation}
In KANs, BlFs serve as interpretable building blocks, designed with the flexibility to be replaced by specific symbolic forms such as polynomial, sine, or logarithmic functions. This design philosophy enables transparent construction of complex functions from simpler components, iterative refinement of the target function during the training process, and potential for direct translation into human-readable mathematical expressions. This modular approach facilitates a bottom-up understanding of the learned function, allowing researchers to dissect and analyze the contribution of each component to the overall model behavior.

HKANs, on the other hand, employ BlFs in a fundamentally different manner. Each BlF attempts to approximate the target function within a one-dimensional space. This approach provides a direct measure of individual input variable importance, with the quality of these one-dimensional approximations serving as a metric for assessing the expressive power of each input variable. 
This characteristic of HKANs allows us to quickly identify key input arguments and gauge their significance in the model, providing a clear path for understanding the contributions of individual variables to the overall function approximation. 

{Other aspects of HKAN explainability are discussed in Section \ref{Disc}. Among these, a particularly valuable feature is the ability to evaluate and interpret the quality of approximation at each stage of hierarchical processing (see Figs. \ref{fig6a}--\ref{fig6d}).}


\subsection{Learning}

KAN and HKAN employ fundamentally different approaches to training, each with distinct characteristics and implications.

KAN utilizes gradient descent in backpropagation process to train the parameters including the weights of the BaFs, $c$, and the weights of the spline and SiLU combinations. This approach allows for fine-tuning of the network but introduces the challenges associated with iterative gradient-based optimization, such as potential convergence to local optima and sensitivity to initial conditions.

HKAN, on the other hand, determines all parameters, i.e. the weights of all linear regressions combining BaFs and blocks, using the least-squares method. 
The training process is non-iterative and hierarchical, progressing along the network structure. Each layer's weights are determined based on the target function predictions made by the preceding linear regressions, which combine either basis functions or blocks from the previous layer.
Only the weights $c$ of the first-layer blocks are directly determined using input data $\mathbf{x}$, while subsequent linear regressions successively refine the fitted function to better approximate the target.

The deterministic and  non-iterative nature of HKAN's training allows for straightforward estimation of computational complexity (see Section \ref{comp}). For KAN, such estimation is challenging due to the unpredictable number of iterations required in the stochastic training process.

HKAN's layer-wise training potentially offers better scalability for deep architectures, as each layer can be optimized independently. KAN's end-to-end training might face challenges with very deep networks due to issues like vanishing gradients.

Table \ref{tab2} summarizes the key differences between KAN and HKAN, providing a quick reference for comparison.

\begin{table}[!t]
 \caption{Comparison of KAN {\cite{Liu24}} and HKAN.}
 \label{tab2}
 \centering
 \begin{tabular}{lll}
\hline
\textbf{Basis functions $g$} \\
\hspace{5pt} KAN: B-splines (order 3)  \\
\hspace{5pt} HKAN: Flexible (e.g., Gaussian, sigmoid) \\
\textbf{Basis function distribution} \\
\hspace{5pt} KAN: Uniform  \\
\hspace{5pt} HKAN: Data-driven or random \\
\textbf{Block functions $\phi$} \\
\hspace{5pt} KAN: Weighted sum of splines (linear combination of BaFs) and SiLU \\
\hspace{5pt} HKAN: Linear combination of BaFs \\
\textbf{Block function combination $h$} \\
\hspace{5pt} KAN: Added without weights \\
\hspace{5pt} HKAN: Linear combination with weights \\
\textbf{Training} \\
\hspace{5pt} KAN: Iterative, gradient-based (backpropagation) \\
\hspace{5pt} HKAN: Non-iterative, hierarchical (least-squares method) \\
\textbf{Explainability} \\
\hspace{5pt} KAN: BlFs can represent interpretable component functions \\
\hspace{5pt} HKAN: Input importance can be assessed based on BlFs; {Allows} \\
\hspace{33pt} {insight into approximation quality at each hierarchical level} \\
\hline
\end{tabular}
\end{table}

\section{HKAN as Multi-Stacking Model}
\label{MSM}

Stacking has emerged as a highly effective approach for enhancing the predictive power of machine learning models \cite{Wol92}. It employs a meta-learning algorithm to optimally combine predictions generated by different learners. By combining multiple diverse weak learners, an ensemble can reduce the overall error. 

In the context of HKAN, BlFs serve as the weak learners, while $h$-functions act as the meta-learners. BlFs typically offer a rough nonlinear approximation of the target function within one-dimensional subspaces, providing distinct perspectives on the input data.

Key aspects of ensembling involve two main considerations: how to combine learners and how to generate diversity among them. Diverse weak learners capture various patterns and relationships within the data. This broad coverage helps the ensemble generalize better to new, unseen data, reducing overfitting to the training set and improving model robustness.

In our case, linear regression addresses the combination of learners, while diversity is achieved through the modeling of the target function in one-dimensional subspaces and the randomized distribution of BaFs. Diversity is further controlled by the number of BaFs and the smoothing parameter.

HKAN builds upon BlFs through a stacking approach. The BlFs, each constructed on different projections of individual inputs, are linearly combined to approximate the target function. This combination is performed by multiple stacking $h$-functions. 
Each stacking function integrates a unique set of BlFs, enabling diverse multivariate representations of the target function.

The process extends hierarchically, with the stacking functions from one layer serving as inputs to the next. In each subsequent layer, these stacking functions are nonlinearly transformed by BaFs and then linearly combined to generate new BlFs. These new BlFs are subsequently aggregated to form the stacking functions of the next layer, resulting in a cascade of increasingly complex and abstract representations.
Notably, each layer consists of multiple stacking functions ($h$), enabling a parallelized process within each level. This multi-level, parallel stacking architecture allows HKAN to efficiently capture intricate relationships in the data, leveraging the strengths of stacking across multiple scales simultaneously.








\section{Experimental Study}
\label{ExSt}

In this section, we compare our proposed HKAN with the standard KAN {and MLP} in regression tasks, evaluating their approximation accuracy. The experimental evaluation was performed on a variety of datasets to validate the model's performance across different regression and function approximation scenarios. 

{The experiments were performed on a virtual machine equipped with 32 vCPUs (host: 2×AMD EPYC 7302), 64 GB of RAM, running Debian 12. The software stack included Python 3.12.6, PyTorch 2.7.1, NumPy 2.1.1, and scikit-learn 1.5.2.}

\subsection{Datasets}

The selected datasets include benchmark regression datasets and synthetically generated data designed to emulate complex target functions. They were chosen to evaluate model's ability to generalize across both simple and highly nonlinear relationships.

The synthetic target functions were defined as follows:

\begin{description}
\item[TF1:] $g(\mathbf{x}) = (2x_1-1)(2x_2-1),\, x_1,x_2 \in [0, 1]$

\item[TF2:] $g(\mathbf{x}) = \sum_{i=1}^{2}\sin\left(20\exp x_i\right) x_i^2, \, x_i \in [0, 1]$ 

\item[TF3:] $g(\mathbf{x}) = -{\sum_{i=1}^{2} x_i \sin(\sqrt{|x_i|})}, \, x_i \in [-500, 500]$

\item[TF4:] $g(\mathbf{x}) = 1- \cos \left( 2\pi\sqrt{\sum_{i=1}^{n} x_i^2}\right) - 0.1\sqrt{\sum_{i=1}^{n} x_i^2},\\
\, n=10, x_i \in [-4, 4]$

\item[TF5:] $g(\mathbf{x}) = -{\sum_{i=1}^{n} \sin(x_i) \sin^{20} \left(\frac{ix_i^2}{\pi}   \right)}, \, n=2, 5,\\
x_i \in [0, \pi]$ 
\end{description}

\begin{figure}[!t]
\centering
\includegraphics[width=0.092\textwidth]{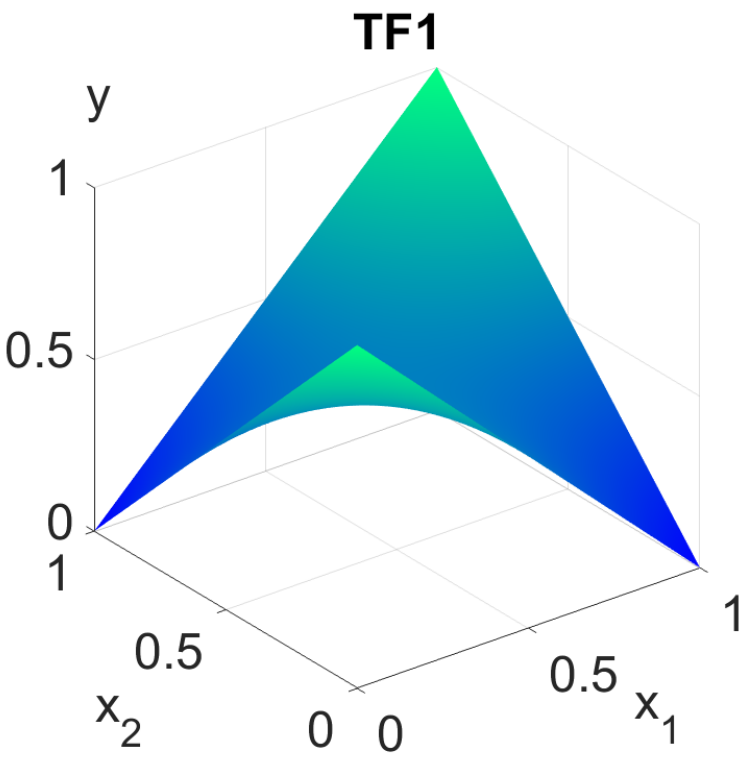}
\includegraphics[width=0.092\textwidth]{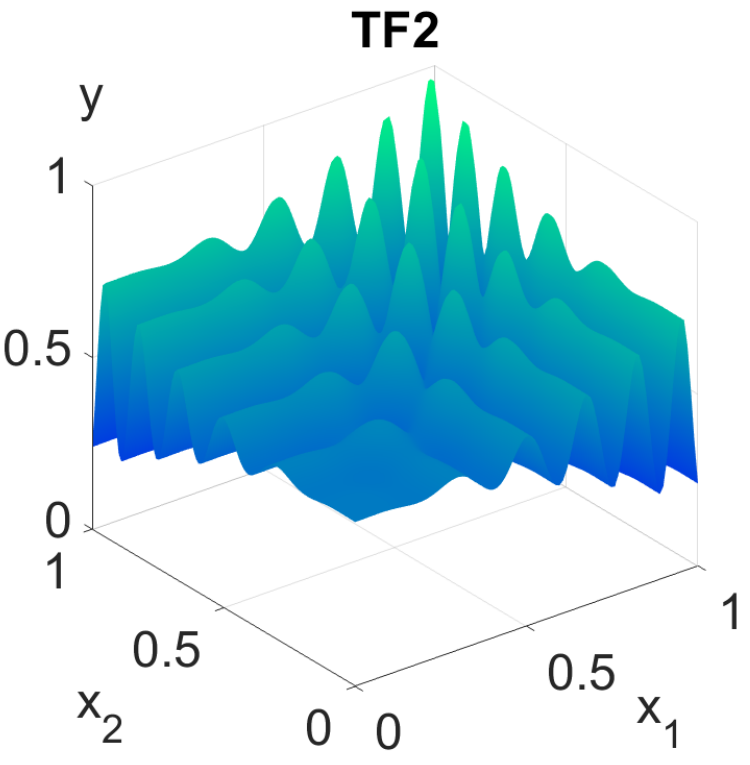}
\includegraphics[width=0.092\textwidth]{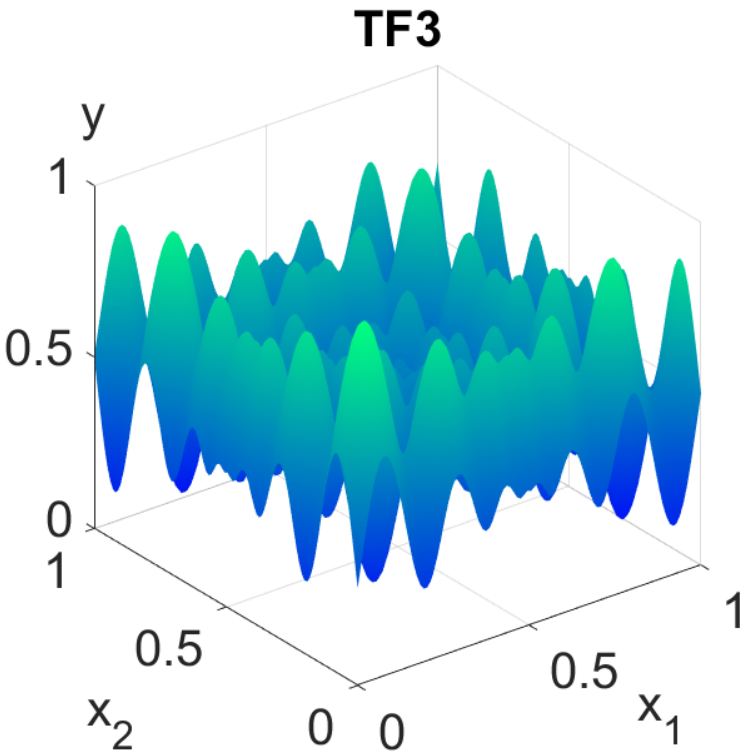}
\includegraphics[width=0.092\textwidth]{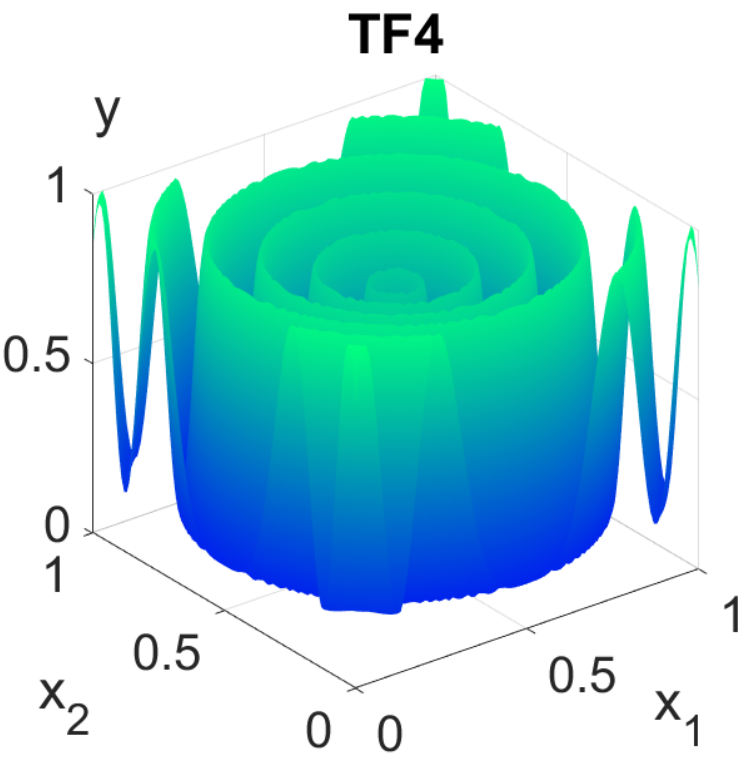}
\includegraphics[width=0.092\textwidth]{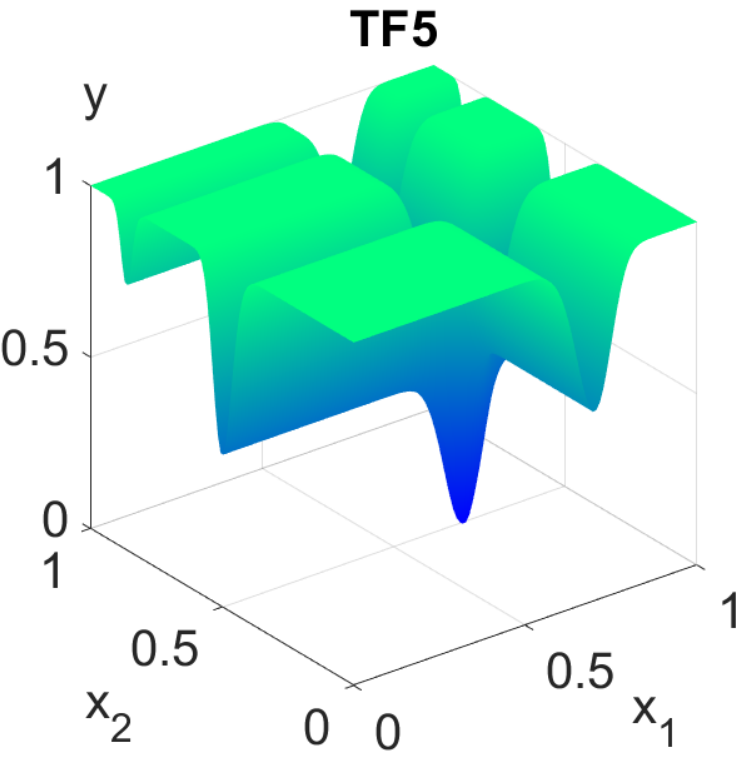}
\caption{Synthetic TFs {(2D variants of TF4 and TF5)}.}
\label{fig3}
\end{figure}

Fig. \ref{fig3} illustrates these functions, each showcasing distinct characteristics. 
TF1 is a simple saddle-shaped function. TF2 is a complex oscillatory function that combines flat regions with strongly fluctuating ones. TF3 is a periodic wave function with consistent oscillations, exhibiting the highest amplitude near the domain borders.
TF4 is a periodic function with radial symmetry. TF5 expresses plateau regions separated by perpendicular grooves of varying depths, with peaks at their intersections.

All functions, except TF4, have two input arguments. TF4 has ten arguments, while TF5 is evaluated both as a two-argument function and a five-argument variant (TF5-5). The function values and input arguments of TF3, TF4, and TF5 were normalized to the range $[0, 1]$. Additionally, the TF2 training data was perturbed by adding noise generated from $U(-0.2, 0.2)$.

Table \ref{tab4} provides an overview of all datasets used in this study, including six synthetically generated datasets and 18 obtained from various sources, as detailed in the table. For these 18 datasets, both input and output variables were normalized to the range $[0, 1]$. The table also specifies the number of samples, input dimensions, and the sizes of the training and test sets. All datasets are available in our GitHub repository \cite{Rod25}.

\subsection{Optimization}

Table \ref{tab3} outlines the search space for HKAN hyperparameters. The tree-structured Parzen estimator algorithm, implemented in the Optuna framework \cite{optuna_2019}, was used to explore this space. A total of 1000 trials were conducted, with early stopping applied by pruning trials where the RMSE exceeded twice the baseline RMSE, calculated as the RMSE of the mean prediction on the training set. The optimal hyperparameters were determined using 5-fold cross-validation. The selected hyperparameter values are summarized in Table \ref{tab3a}.


\begin{table}[!t]
\caption{Hyperparameter search space for HKAN.}
\label{tab3}
\centering
\begin{tabular}{ll}
\hline
{Hyperparameter} & {Search space} \\
\hline
\#layers, $L$ & $\{1, 2, 3\}$ \\
\#nodes, $n^{(l)}$ & $\{2, \ldots, 1000(200)^*\}$\\
BaF type & Sigmoid (S), Gaussian (G), ReLU (R),  \\
& Softplus (S+), Tanh (T), Identity$^{**}$ (I) \\
Smoothing param., $\sigma^{(l)}$ & $\{1, \ldots, 50\}$ \\
\#BaFs, $m^{(l)}$ & $\{1, \ldots, 40\}$ \\
BaF distribution & Random (R), Data-driven (D), \\
& Equally spaced$^{**}$ (E) \\
Regularization param., $\lambda^{(l)}_\phi$  & $\{0, 0.001, 0.01, 0.1, 1, 10\}$ \\
\hline
\end{tabular}
  \vspace{1ex}
 
  {\raggedright $\quad ^*$1000 for 1- and 2-layer nets, 200 for 3-layer nets.\\ $\quad  ^{**}$Only for the output layer.
  \par}
\end{table}

\begin{table}[]
  \caption{Hyperparameters selected for HKAN.}
  \label{tab3a}
  \centering
\setlength{\tabcolsep}{2pt}
\scriptsize
\begin{tabular}{l|l|l|l|l|l|l|l}
\hline
Data & \multicolumn{1}{c|}{$L$} & \multicolumn{1}{c|}{$n^*$} & \multicolumn{1}{c|}{BaF type$^{**}$} & \multicolumn{1}{c|}{$\sigma$} & \multicolumn{1}{c|}{$m$} & \multicolumn{1}{c|}{BaF distr.} & \multicolumn{1}{c}{$\lambda$} \\
\hline
TF1   & 2   & 932  & S, T & 1, 33 & 2, 13  & D, D   & 0.1, 10  \\
TF2   & 3   & 48, 11 & T, S+, I & 22, 18& 17, 21 & R, D & 0.1, 1  \\
TF3   & 2   & 924  & T, I & 50 & 39   & R & .001 \\
TF4   & 1   &  & T& 3  & 2 & E & 0.001   \\
TF5   & 2   & 912  & T, I & 50 & 23   & R & 0.01   \\
TF5-5 & 2   & 1000 & T, I & 30 & 32   & R   & 0.001 \\
Abal. & 1   &  & I &  &  &   & \\
Auto. & 1   &  & S & 9  & 32   & E & 0.001  \\
Bank. & 1   &  & I &  &  &   & \\
Comp. & 3   & 189, 30   & T, S, R  & 4, 3, 44   & 2, 38, 9  & D, D, R & 10, 1, 0.01 \\
Conc. & 3   & 153, 23   & S, S, S  & 32, 3, 19  & 10, 2, 11 & D, R, D & 0.01, 1, 0.01   \\
Dee   & 1   &  & S & 2  & 38   & D & 0.1 \\
Ele2  & 2   & 478  & R, I & 19 & 39   & D& 0.01   \\
Elev. & 2   & 926  & R, S & 22, 6 & 1, 11  & D, R   & 0.001, 0.1  \\
Kin8  & 3   & 200, 171  & G, S, R  & 1, 5, 27   & 4, 3, 30  & D, R, R & 1, 0.01, 1 \\
Kin32 & 1   &  & S+   & 1  & 40   & E & 0.001  \\
Laser & 2   & 86   & S, S+ & 23, 15& 1, 35  & D, D   & 1, 10  \\
Mach. & 1   &  & R & 1  & 25   & E & 1  \\
Puma. & 2   & 628  & G, S+ & 2, 1  & 1, 12  & D, D   & 0.001, 0.001\\
Pyra. & 2   & 678  & G, T & 48, 33& 2, 20  & R, D   & 10, 10 \\
Stock & 3   & 197, 72   & S, S+, S & 15, 37, 42 & 1, 3, 24  & R, D, R & 0.001, 0.001, 0.1   \\
Treas.& 2   & 97   & S+, I & 3  & 35   & D & 1 \\
Triaz.& 2   & 2 & S, G & 40, 41& 29, 6  & R, D   & 1, 0.01  \\
Wiz.  & 1   &  & I &  &  &   &\\
\hline
\end{tabular}
   \vspace{1ex}

{\raggedright $ ^{*}$For the final layer, $n^{(L)}=1$ (not shown).\\
$ ^{**}$The identity function (I) does not require any parameters ($\sigma$, $m$, $\lambda_\phi$, and BaF distribution).
 \par}
\end{table}

KAN optimization was performed using Optuna's trial system, integrated with a grid search sampler \cite{optuna_2019}.
The network’s architecture was optimized across a predefined set of configurations:
 $\mathcal{W} = \{[n, 1],[n, 2, 1],[n, n + 1, 1],[n, 2n + 1, 1],
[n, 2, 2, 1],[n, n + 1, 2, 1],[n, 2n + 1, 2, 1],
[n, n + 1, n + 1, 1],[n, 2n + 1, n + 1, 1], [n, 2n + 1, 2n + 1, 1]\}$. 
These architectures were chosen based on the original authors’ recommendations, emphasizing model accuracy over interpretability. 


Initial experiments highlighted the importance of identifying the optimal number of training steps to maximize the performance of the KAN model, as excessive training can lead to overfitting. To mitigate this risk, a 5-fold cross-validation approach was employed to determine the ideal number of training steps at each grid resolution level within a multi-resolution framework. This process was guided by a stopping criterion that halts training when further grid refinement no longer improves model performance.

In summary, the KAN hyperparameter optimization process involved the following steps:
\begin{enumerate}
 \item Iteratively identifying the optimal number of training steps for each grid resolution for every KAN architecture in $\mathcal{W}$.
 \item Evaluating model performance through cross-validation using these optimal training steps.
 \item Selecting the architecture with the best cross-validation performance.
\end{enumerate}

{MLP optimization was carried out using Optuna with 2000 trials. The optimized hyperparameters included depth, hidden size, activation function type, learning rate, batch size, dropout rate, and patience. The search ranges for these hyperparameters were adjusted according to the size of the training datasets. For example, for smaller datasets the maximum depth was limited to 2 and the hidden size to 64, whereas for the largest datasets the depth was increased to 5 and the hidden size to 256.}   

\subsection{Results}

\begin{figure}[!t]
\centering
\includegraphics[width=0.48\textwidth]{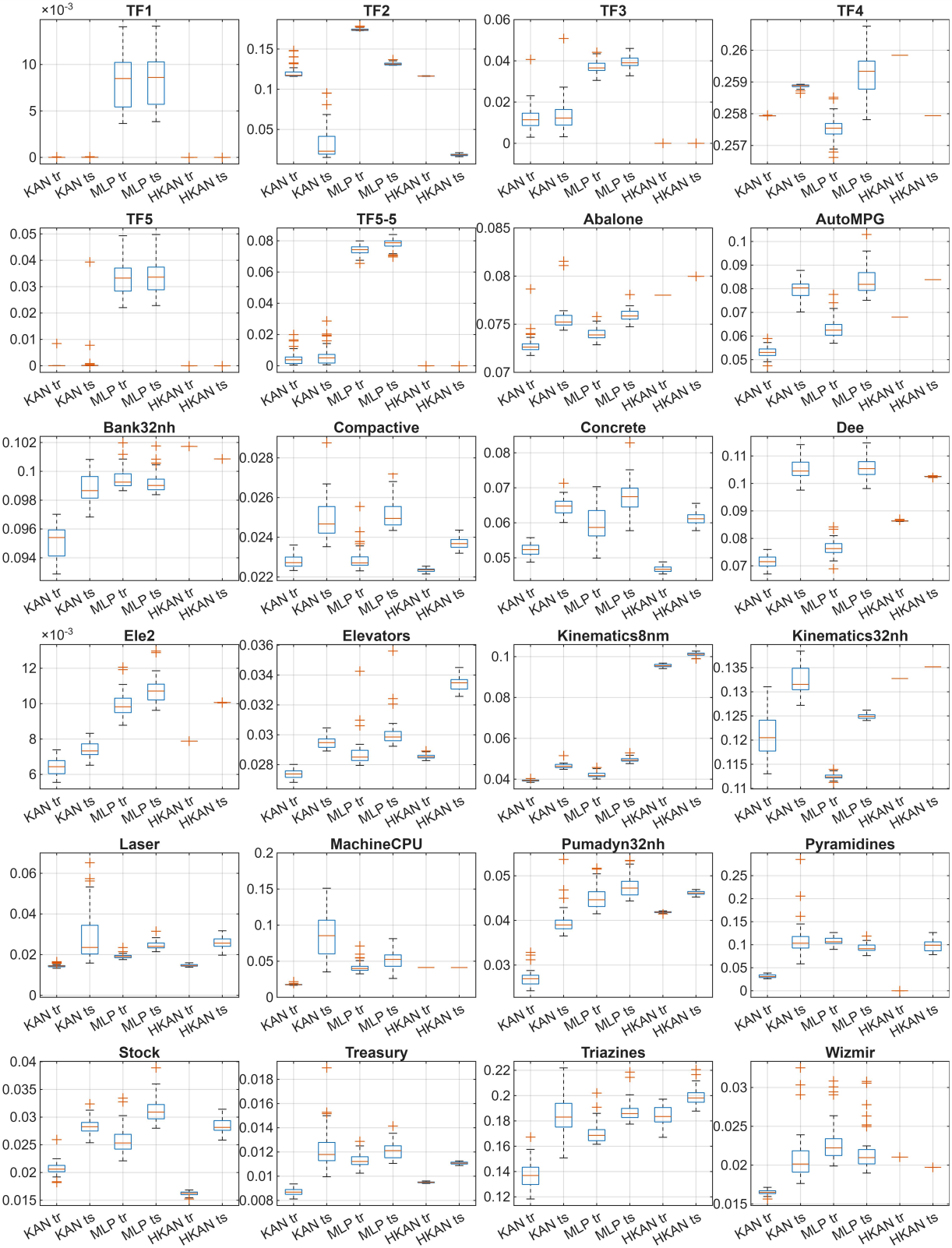}
\caption{{Distribution of training (tr) and test (ts) RMSE for KAN, MLP and HKAN.}}
\label{fig6}
\end{figure}

Table \ref{tab4} summarizes the performance metrics for KAN and HKAN, including the median and interquartile range (IQR) of RMSE for both training and test data, calculated from 50 independent training sessions per model. These results are further illustrated in Fig. \ref{fig6} using boxplots.

\begin{table*}[t]
  \caption{Performance comparison of KAN and HKAN.}
  \label{tab4}
  \centering
\begin{tabular}{l|l|r|r|r|r|r|r|r|r}
\hline
   &  & \multicolumn{4}{c|}{KAN} & \multicolumn{4}{c}{HKAN}   \\ \cline{3-10} 
Data & \#samples (training/test) /   & \multicolumn{2}{c|}{Training RMSE} & \multicolumn{2}{c|}{Test RMSE}  & \multicolumn{2}{c|}{Training RMSE} & \multicolumn{2}{c}{Test RMSE}  \\ \cline{3-10} 
   & \#arguments & \multicolumn{1}{c|}{Median} & \multicolumn{1}{c|}{IQR} & \multicolumn{1}{c|}{Median} & \multicolumn{1}{c|}{IQR} & \multicolumn{1}{c|}{Median} & \multicolumn{1}{c|}{IQR} & \multicolumn{1}{c|}{Median} & \multicolumn{1}{c}{IQR} \\ \hline
TF1  & 15000 (5000/10000) / 2  & { 5.15E-06}  & 5.59E-06   & 1.09E-05  & 1.27E-05   & { 2.68E-14}  & 9.62E-15   & \textbf{3.94E-14}  & 2.00E-14  \\
TF2  & 15000 (5000/10000) / 2  & 1.17E-01  & 4.82E-03   & {\ul 2.29E-02}  & 2.21E-02   & { 1.16E-01}  & 2.21E-04   &{\ul \textbf{1.85E-02}}  & 1.64E-03  \\
TF3  & 15000 (5000/10000) / 2  & { 1.14E-02}  & 6.00E-03   & 1.22E-02  & 7.58E-03   & 6.37E-06  & 5.88E-08   & {\ul \textbf{5.08E-06}} & 3.85E-07  \\
TF4  & 5000 (3750/1250) / 10  & { 2.58E-01}  & 4.50E-06   & 2.59E-01  & 5.42E-05   & 2.60E-01  & 0 & {\ul \textbf{2.58E-01}} & 0\\
TF5  & 15000 (5000/10000) / 2  & { 3.31E-05}  & 3.26E-05   & 8.58E-05  & 1.06E-04   & { 2.98E-15}  & 4.58E-16   & \textbf{4.56E-15}  & 2.39E-15  \\
TF5-5   & 10000 (7500/2500) / 5   & { 3.83E-03}  & 4.12E-03   & 5.07E-03  & 5.51E-03   & { 3.40E-09}  & 8.70E-10   & \textbf{3.93E-09}  & 9.88E-10  \\
Abalone \cite{Bac17} & 4177 (3133/1044) / 8 & 7.26E-02  & 6.00E-04   & \textbf{7.52E-02}  & 1.00E-03   & 7.80E-02  & 0   & 8.00E-02  & 0\\
AutoMPG \cite{Bac17} & 386 (270/116) / 7   & 5.31E-02  & 2.78E-03   & \textbf{8.04E-02}  & 4.85E-03   & 6.80E-02  & 0 & 8.38E-02  & 0\\
Bank32nh \cite{Del16}   & 8192 (5734/2458) / 32   & { 9.54E-02}  & 1.81E-03   & \textbf{9.87E-02}  & 1.50E-03   & 1.02E-01  & 0 & {\ul 1.01E-01}  & 0\\
Compactive \cite{Alc11} & 8192 (6144/2048) / 21   & { 2.27E-02}  & 4.61E-04   & 2.47E-02  & 1.34E-03   & { 2.24E-02}  & 1.19E-04   & \textbf{2.37E-02}  & 3.94E-04  \\
Concrete \cite{Bac17}   & 1030 (773/257) / 8  & 5.23E-02  & 2.55E-03   & 6.48E-02  & 3.33E-03   & 4.67E-02  & 1.26E-03   & \textbf{6.12E-02}  & 2.30E-03  \\
Dee \cite{Alc11}   & 365 (274/91) / 6 & 7.15E-02  & 3.29E-03   & 1.05E-01  & 4.85E-03   & { 8.63E-02}  & 1.10E-04   & \textbf{1.03E-01}  & 1.22E-04  \\
Ele2 \cite{Alc11}  & 1056 (792/264) / 4  & { 6.44E-03}  & 7.37E-04   & \textbf{7.33E-03}  & 6.07E-04   & { 7.88E-03}  & 4.78E-07   & 1.01E-02  & 9.85E-07  \\
Elevators \cite{Alc11}  & 16599 (11619/4980) / 18 & { 2.74E-02}  & 4.37E-04   & \textbf{2.95E-02}  & 5.62E-04   & 2.85E-02  & 1.73E-04   & 3.35E-02  & 6.34E-04  \\
Kinematics8nm \cite{Del16}  & 8192 (6144/2048) / 8 & 3.94E-02  & 4.77E-04   & \textbf{4.63E-02}  & 1.57E-03   & { 9.55E-02}  & 1.15E-03   & 1.01E-01  & 1.09E-03  \\
Kinematics32nh \cite{Del16} & 8192 (5734/2458) / 32   & { 1.21E-01}  & 6.38E-03   & \textbf{1.32E-01}  & 4.44E-03   & { 1.33E-01}  & 0 & 1.35E-01  & 0\\
Laser \cite{Alc11} & 993 (745/248) / 4   & { 1.44E-02}  & 5.78E-04   & 2.35E-02  & 1.41E-02   & { 1.46E-02}  & 7.41E-04   & 2.56E-02  & 3.62E-03  \\
MachineCPU \cite{Alc11} & 209 (146/63) / 6 & { 1.76E-02}  & 6.03E-04   & 8.54E-02  & 4.66E-02   & 4.13E-02  & 0 & \textbf{4.13E-02}  & 0\\
Pumadyn32nh \cite{Del16} & 8192 (5734/2458) / 32   & { 2.69E-02}  & 1.95E-03   & \textbf{3.90E-02}  & 1.93E-03   & 4.19E-02  & 2.08E-04   & 4.61E-02  & 6.03E-04  \\
Pyramidines \cite{Tor17} & 74 (52/22) / 27& { 3.08E-02}  & 5.72E-03   & 1.03E-01  & 2.50E-02   & { 7.34E-16}  & 1.40E-16   & 9.89E-02  & 1.93E-02  \\
Stock \cite{Alc11} & 950 (713/237) / 9   & { 2.06E-02}  & 1.14E-03   & 2.83E-02  & 1.55E-03   & { 1.63E-02}  & 4.77E-04   & 2.81E-02  & 1.75E-03  \\
Treasury \cite{Alc11}   & 1049 (734/315) / 15 & { 8.69E-03}  & 3.99E-04   & 1.18E-02  & 1.50E-03   & { 9.49E-03}  & 6.41E-05   & \textbf{1.11E-02}  & 1.56E-04  \\
Triazines \cite{Tor17}  & 186 (130/56) / 60   & { 1.37E-01}  & 1.37E-02   & \textbf{1.83E-01}  & 1.88E-02   & { 1.83E-01}  & 1.14E-02   & 1.98E-01  & 7.51E-03  \\
Wizmir \cite{Alc11}& 1461 (1096/365) / 9 & { 1.65E-02}  & 3.60E-04   & 2.01E-02  & 2.75E-03   & 2.10E-02  & 0 & {\ul \textbf{1.97E-02}} & 0  \\ \hline
 
  \end{tabular}
   \vspace{1ex}

  {\raggedright The test errors of both models were compared using the Wilcoxon test, with significantly lower values highlighted in \textbf{bold}. \\
  The test and training errors were compared separately for each model using the Wilcoxon test, with significantly lower test errors {\ul underlined}.
  \par}

\end{table*}

The test errors of both models were compared using the Wilcoxon signed-rank test. As shown in Table \ref{tab4}, HKAN achieved significantly lower errors than KAN in 12 out of 24 cases, while KAN outperformed HKAN in 9 cases. Notably, HKAN demonstrated superior accuracy on synthetic functions TF1, TF3, TF5, and TF5-5, where its accuracy exceeded that of KAN by several orders of magnitude. For TF2 (a synthetic function with noise), HKAN’s errors were over 19\% lower than KAN’s. A substantial improvement was also observed for the MachineCPU dataset, where HKAN outperformed KAN with a difference exceeding 51\%.

Conversely, the largest differences favoring KAN were observed for the Kinematics8nm dataset (over 118\%), Ele2 (over 37\%), Pumadyn32nh (over 18\%), and Elevators (over 13\%). For the remaining datasets, the differences in test errors between the two models were within 10\%.

When comparing training and test errors for each model separately, it was observed that, in most cases, the training error was significantly lower than the test error, as confirmed by the Wilcoxon test. Cases where the test error was significantly lower than the training error were rare, occurring five times for HKAN and only once for KAN (see the underlined errors in Table \ref{tab4}). Substantially lower training errors compared to test errors may indicate overfitting, differences in the distributions of training and test sets, insufficient representation of the test set in the training data, or an inadequate number of training samples relative to the number of input arguments.

The IQR serves as a measure of model variance, reflecting the consistency of predictions across different training sessions. As shown in Table \ref{tab4} and Fig. \ref{fig6}, HKAN produces more stable results than KAN in all cases except for the Elevators and Stock datasets. HKAN frequently exhibits an IQR of 0, indicating deterministic behavior. This occurs when the optimal architecture comprises a single layer ($L=1$) with identity BaFs or uniformly spaced BaFs (refer to Table \ref{tab3a} for the optimal hyperparameters). Under such conditions, HKAN consistently produces identical results across all training sessions.

{Table \ref{tab4a} compares HKAN with MLP. In this table, the cases where one model’s test error is significantly lower than the competitor’s are highlighted in bold. There are 15 such cases in favor of HKAN and 8 in favor of MLP. Of particular note are the substantially lower errors achieved by HKAN on synthetic datasets, with the exception of TF4. For the remaining datasets, the differences in test errors did not exceed 10\%, except for MachineCPU (approximately 22\% in favor of HKAN), Kinematics8nm (approximately 105\% in favor of MLP), and Elevators (approximately 12\% in favor of MLP).}

{It is also worth noting that the variance of errors in the case of HKAN is significantly smaller than for MLP. Only in four cases does MLP exhibit lower variance than HKAN (see Fig. \ref{fig6} and the IQR values in Table \ref{tab4a}). This indicates that HKAN provides more consistent and stable predictions across independent training sessions.}

\begin{table*}[t]
  \caption{{Performance comparison of MLP and HKAN.}}
  \setlength{\tabcolsep}{12pt}
  \label{tab4a}
  \centering
\begin{tabular}{l|r|r|r|r|r|r|r|r}
\hline
 & \multicolumn{4}{c|}{MLP}& \multicolumn{4}{c}{HKAN}   \\ \cline{2-9} 
Data & \multicolumn{2}{c|}{Training RMSE} & \multicolumn{2}{c|}{Test RMSE}  & \multicolumn{2}{c|}{Training RMSE} & \multicolumn{2}{c}{Test RMSE}  \\ \cline{2-9} 
   & \multicolumn{1}{c|}{Median} & \multicolumn{1}{c|}{IQR} & \multicolumn{1}{c|}{Median} & \multicolumn{1}{c|}{IQR} & \multicolumn{1}{c|}{Median} & \multicolumn{1}{c|}{IQR} & \multicolumn{1}{c|}{Median} & \multicolumn{1}{c}{IQR} \\ \hline
TF1  & 8.49E-03 & 4.80E-03 & 8.60E-03 & 4.56E-03 & 2.68E-14 & 9.62E-15 & \textbf{3.94E-14}    & 2.00E-14 \\
TF2  & 1.74E-01 & 1.72E-03 & {\ul 1.31E-01}   & 2.27E-03 & 1.16E-01 & 2.21E-04 & {\ul \textbf{1.85E-02}} & 1.64E-03 \\
TF3  & 3.66E-02 & 3.50E-03 & 3.90E-02 & 3.59E-03 & 6.37E-06 & 5.88E-08 & {\ul \textbf{5.08E-06}} & 3.85E-07 \\
TF4  & 2.58E-01 & 3.31E-04 & 2.59E-01 & 8.85E-04 & 2.60E-01 & 0 & {\ul \textbf{2.58E-01}} & 0 \\
TF5  & 3.33E-02 & 8.74E-03 & 3.36E-02 & 8.60E-03 & 2.98E-15 & 4.58E-16 & \textbf{4.56E-15}    & 2.39E-15 \\
TF5-5   & 7.43E-02 & 3.78E-03 & 7.87E-02 & 3.33E-03 & 3.40E-09 & 8.70E-10 & \textbf{3.93E-09}    & 9.88E-10 \\
Abalone & 7.39E-02 & 7.71E-04 & \textbf{7.59E-02}    & 8.08E-04 & 7.80E-02 & 1.39E-17 & 8.00E-02 & 0 \\
AutoMPG & 6.25E-02 & 4.54E-03 & 8.19E-02 & 7.53E-03 & 6.80E-02 & 0 & 8.38E-02 & 0 \\
Bank32nh    & 9.93E-02 & 8.24E-04 & {\ul \textbf{9.90E-02}} & 7.36E-04 & 1.02E-01 & 0 & {\ul 1.01E-01}   & 0 \\
Compactive  & 2.27E-02 & 4.23E-04 & 2.50E-02 & 9.26E-04 & 2.24E-02 & 1.19E-04 & \textbf{2.37E-02}    & 3.94E-04 \\
Concrete    & 5.87E-02 & 7.25E-03 & 6.75E-02 & 5.38E-03 & 4.67E-02 & 1.26E-03 & \textbf{6.12E-02}    & 2.30E-03 \\
Dee  & 7.63E-02 & 3.24E-03 & 1.05E-01 & 4.65E-03 & 8.63E-02 & 1.10E-04 & \textbf{1.03E-01}    & 1.22E-04 \\
Ele2 & 9.82E-03 & 8.15E-04 & 1.07E-02 & 8.93E-04 & 7.88E-03 & 4.78E-07 & \textbf{1.01E-02}    & 9.85E-07 \\
Elevators   & 2.85E-02 & 6.85E-04 & \textbf{2.99E-02}    & 6.18E-04 & 2.85E-02 & 1.73E-04 & 3.35E-02 & 6.34E-04 \\
Kinematics8nm  & 4.17E-02 & 1.62E-03 & \textbf{4.93E-02}    & 1.22E-03 & 9.55E-02 & 1.15E-03 & 1.01E-01 & 1.09E-03 \\
Kinematics32nh & 1.12E-01 & 5.98E-04 & \textbf{1.25E-01}    & 7.40E-04 & 1.33E-01 & 0 & 1.35E-01 & 0 \\
Laser   & 1.91E-02 & 1.14E-03 & \textbf{2.40E-02} & 2.36E-03 & 1.46E-02 & 7.41E-04 & 2.56E-02 & 3.62E-03 \\
MachineCPU  & 4.00E-02 & 5.56E-03 & 5.26E-02 & 1.57E-02 & 4.13E-02 & 0 & \textbf{4.13E-02}    & 0 \\
Pumadyn32nh & 4.46E-02 & 3.32E-03 & 4.72E-02 & 3.05E-03 & 4.19E-02 & 2.08E-04 & \textbf{4.61E-02}    & 6.03E-04 \\
Pyramidines & 1.06E-01 & 1.11E-02 & {\ul \textbf{9.14E-02}} & 1.15E-02 & 7.34E-16 & 1.40E-16 & 9.89E-02 & 1.93E-02 \\
Stock   & 2.53E-02 & 2.67E-03 & 3.09E-02 & 2.58E-03 & 1.63E-02 & 4.77E-04 & \textbf{2.81E-02}    & 1.75E-03 \\
Treasury    & 1.12E-02 & 6.41E-04 & 1.21E-02 & 9.85E-04 & 9.49E-03 & 6.41E-05 & \textbf{1.11E-02}    & 1.56E-04 \\
Triazines   & 1.69E-01 & 8.73E-03 & \textbf{1.86E-01}    & 7.20E-03 & 1.83E-01 & 1.14E-02 & 1.98E-01 & 7.51E-03 \\
Wizmir  & 2.22E-02 & 2.17E-03 & {\ul 2.10E-02}   & 1.85E-03 & 2.10E-02 & 0 & {\ul \textbf{1.97E-02}} & 1.73E-17\\ \hline

  \end{tabular}
   \vspace{1ex}

  {\raggedright The test errors of both models were compared using the Wilcoxon test, with significantly lower values highlighted in \textbf{bold}. \\
  The test and training errors were compared separately for each model using the Wilcoxon test, with significantly lower test errors {\ul underlined}.
  \par}

\end{table*}

{Table \ref{tab4b} presents the average training times for the optimal configurations of KAN, MLP, and HKAN. It is worth noting that HKAN achieved the shortest training time for 17 datasets, KAN for 4, and MLP for 3.}

\begin{table}[t]
  \caption{{Average training time in seconds.}}
  \setlength{\tabcolsep}{15pt}
  \label{tab4b}
  \centering
  \begin{tabular}{lrrr}
  \hline
  Data           & \multicolumn{1}{c}{KAN} & \multicolumn{1}{c}{MLP} & \multicolumn{1}{c}{HKAN} \\
  \hline
TF1            & 113.09   & 36.91    & \textbf{8.80}            \\
TF2            & 74.37    & 8.88     & \textbf{6.44}            \\
TF3            & 333.50   & 48.18    & \textbf{38.75}           \\
TF4            & 0.56     & 3.88     & \textbf{0.02}            \\
TF5            & 553.88   & 27.12    & \textbf{23.17}           \\
TF5-5          & 633.43   & \textbf{93.27}          & 100.62    \\
Abalone        & 3.86     & 4.46     & \textbf{0.01}            \\
AutoMPG        & 0.62     & 0.39     & \textbf{0.01}            \\
Bank32nh       & 2.38     & 14.33    & \textbf{0.03}            \\
Compactive     & \textbf{28.17}          & 86.18    & 68.69     \\
Concrete       & \textbf{2.20}           & 1.27     & 3.95      \\
Dee            & 0.44     & 0.47     & \textbf{0.02}            \\
Ele2           & 38.43    & 4.20     & \textbf{3.25}            \\
Elevators      & 69.18    & 90.33    & \textbf{24.37}           \\
Kinematics8nm  & 82.42    & 67.03    & \textbf{58.04}           \\
Kinematics32nh & 12.94    & 34.67    & \textbf{0.49}            \\
Laser          & 11.95    & 2.82     & \textbf{0.52}            \\
MachineCPU     & 0.96     & 0.16     & \textbf{0.01}            \\
Pumadyn32nh    & 71.70    & 64.01    & \textbf{23.17}           \\
Pyramidines    & \textbf{0.17}           & 0.11     & 12.23     \\
Stock          & \textbf{4.48}           & 5.57     & 13.34     \\
Treasury       & 8.25     & 3.64     & \textbf{3.50}            \\
Triazines      & 4.48     & \textbf{0.13}           & 13.34     \\
Wizmir         & 8.25     & \textbf{1.32}           & 3.50
\\ \hline
  \end{tabular}
\end{table}

\subsection{How HKAN Constructs Fitted Function}

\begin{figure}[!t]
\centering
\includegraphics[width=0.49\textwidth]{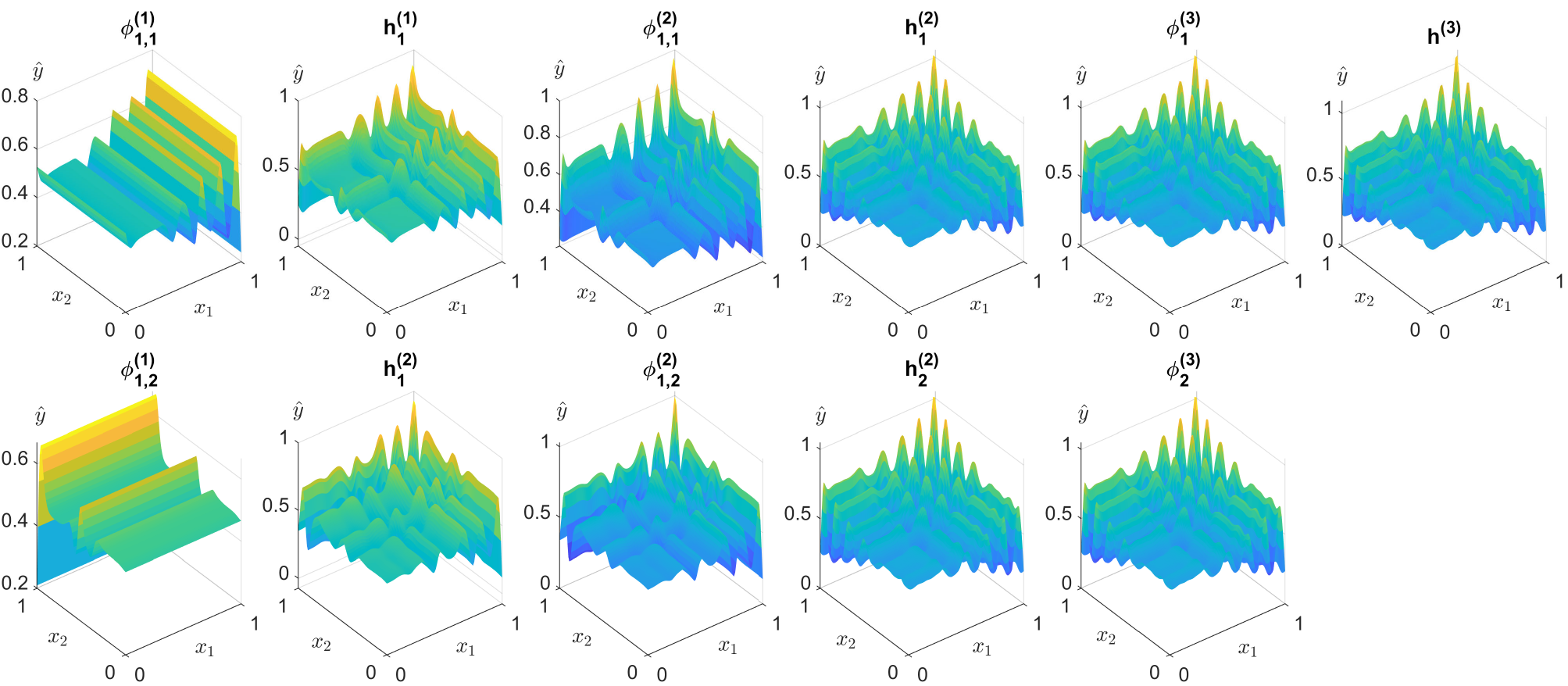}
\includegraphics[width=0.49\textwidth]{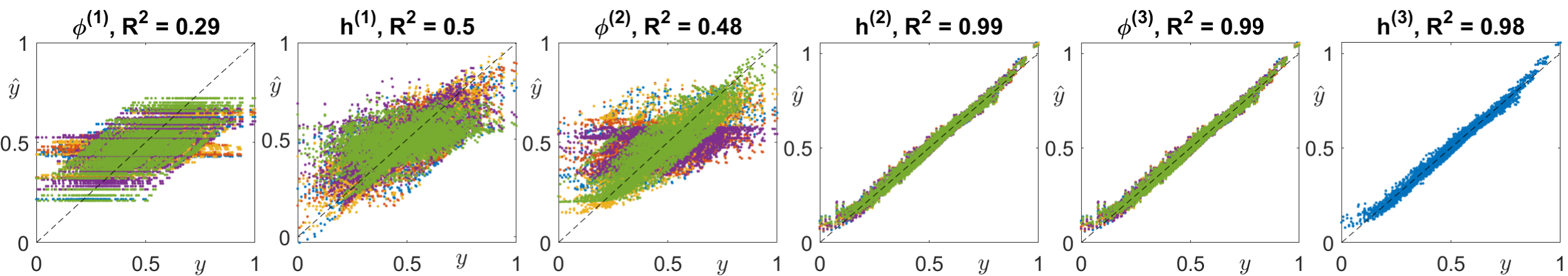}
\caption{Fitted functions and predicted vs. target plots at successive levels of HKAN processing for TF2.}
\label{fig6a}
\end{figure}

This section analyzes how HKAN constructs a fitted function layer by layer. Fig. \ref{fig6a} provides an example for TF2. The two upper panels illustrate the functions fitted at successive levels of HKAN processing (successive linear regressions), specifically the BlFs of the first layer ($\phi^{(1)}$), the $h$-functions of the first layer ($h^{(1)}$), the BlFs of the second layer ($\phi^{(2)}$), the $h$-functions of the second layer ($h^{(2)}$), the BlFs of the third layer ($\phi^{(3)}$), and the $h$-function of the third layer ($h^{(3)}$). 

At each level, except the final one, multiple functions are fitted in parallel; for clarity, only two representative functions are displayed in the two upper panels. The lower panel presents predicted vs. target plots for five selected fitted functions at each processing level, excluding the final level, where only a single function is created.

The following insights can be drawn from Fig. \ref{fig6a}:
\begin{enumerate}
 \item Shapes and Complexity of the Fitted Functions: 
 The fitted function evolves in shape and complexity as it progresses through the layers. At the first level, the target function is modeled nonlinearly using individual input variables, capturing only the features apparent in these variables, such as localized fluctuations. The second level integrates these preliminary approximations across all input variables, producing a multi-variable approximation that remains relatively coarse. The subsequent two levels -- nonlinear transformations by the blocks of the second layer followed by linear combinations of their outputs -- significantly improve the approximation quality.
 The fifth and sixth levels refine the result by processing the fourth-level outputs in a similar manner.

 \item Modeling Variance: 
 The modeling variance, represented by deviations from the diagonal zero-error line in the predicted vs. target plots, decreases significantly across successive levels.
 At the first level, the fitted functions display high variance and are constrained to a narrow range of approximately 0.2-0.75.
In subsequent levels, the range progressively expands. By the forth level, the modeling variance is significantly reduced, and the range widens to its full extent. However, small deviations from the diagonal persist at the boundaries, indicating that the extreme values of the target function are not fully captured by the HKAN model.

\item Diversity in Blocks and Nodes: 
Each block produces a distinct BlF due to differing inputs and the distribution of BaFs. Significant diversity is observed among the BlFs in the first and second layers, as well as among the nodes in the first layer. However, this diversity diminishes at the second layer’s output, where individual nodes achieve a more accurate approximation of the target function. In the third layer, diversity among BlFs is limited, as blocks in this layer process more uniform inputs.

\end{enumerate}

\begin{figure}[!t]
\centering
\includegraphics[width=0.48\textwidth]{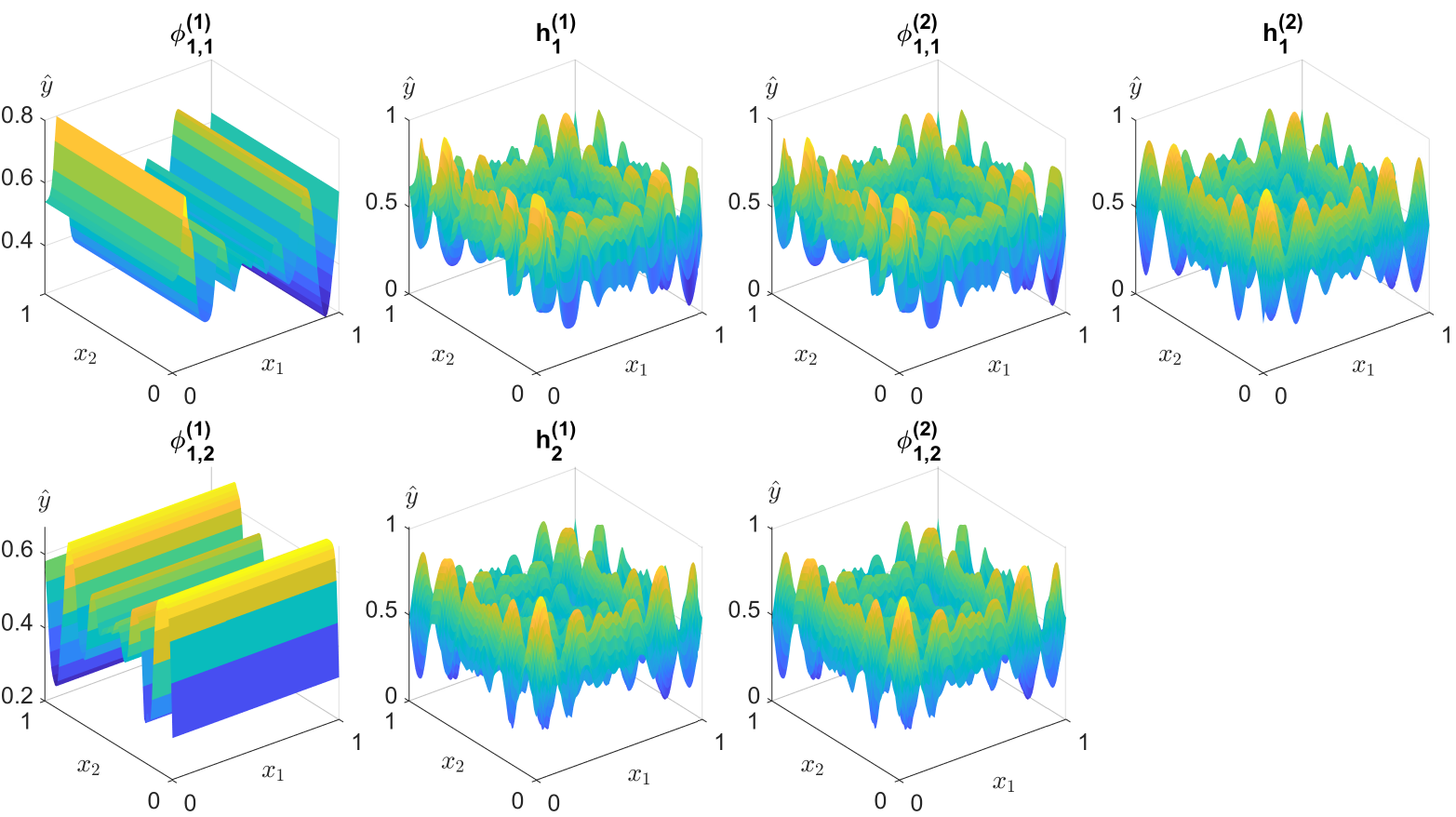}
\includegraphics[width=0.48\textwidth]{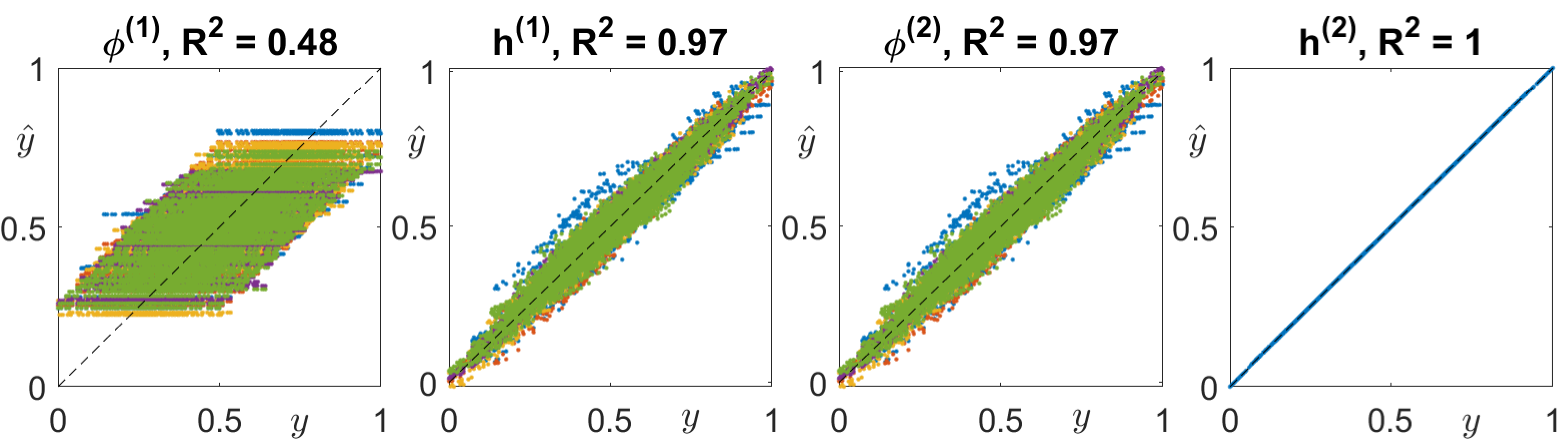}
\caption{Fitted functions and predicted vs. target plots at successive levels of HKAN processing for TF3.}
\label{fig6b}
\end{figure}

\begin{figure}[!t]
\centering
\includegraphics[width=0.48\textwidth]{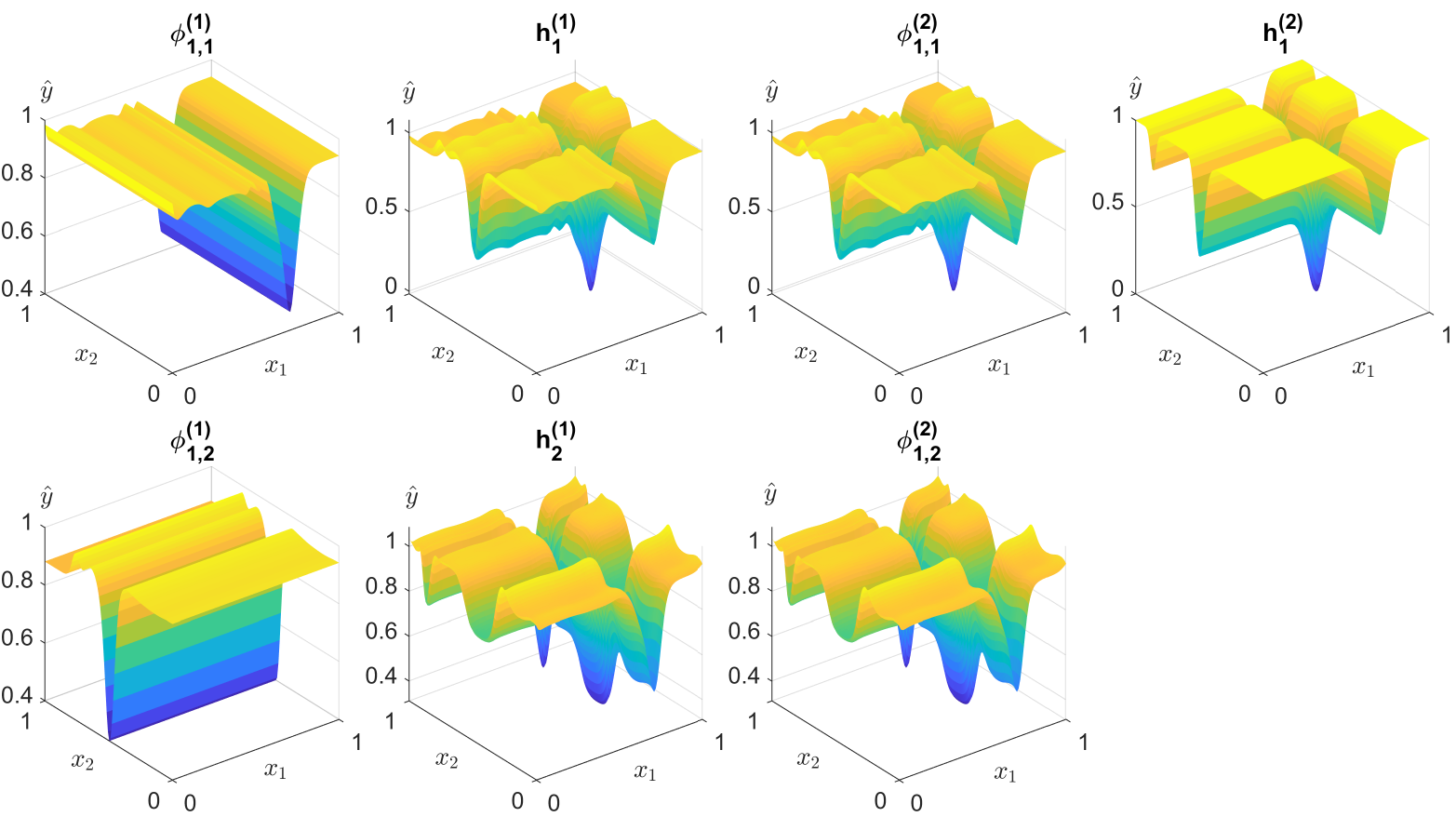}
\includegraphics[width=0.48\textwidth]{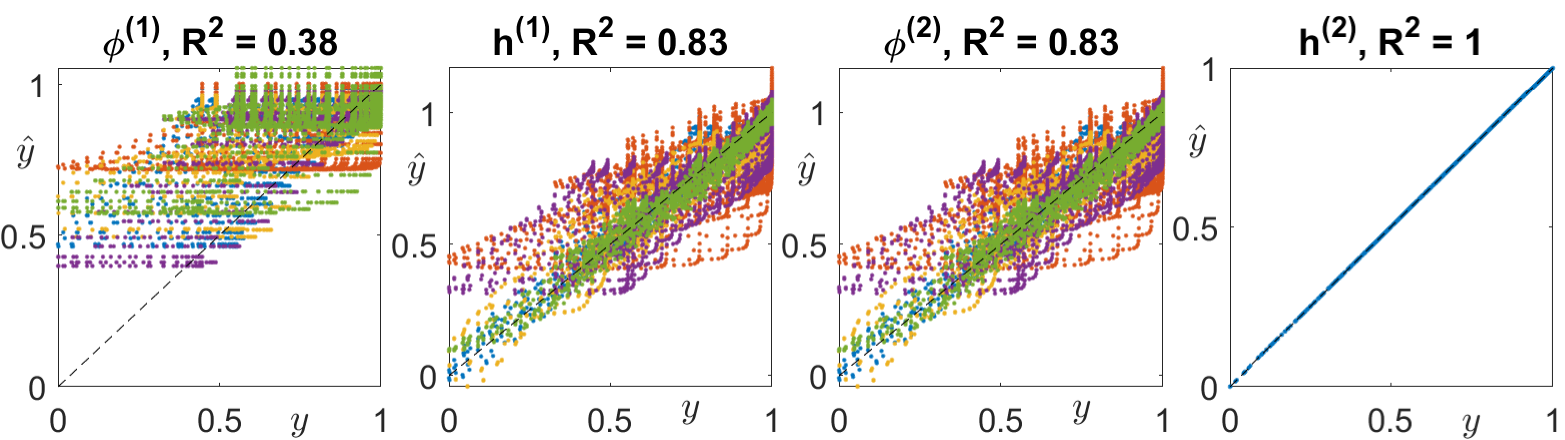}
\caption{Fitted functions and predicted vs. target plots at successive levels of HKAN processing for TF5.}
\label{fig6c}
\end{figure}

Figs. \ref{fig6b} and \ref{fig6c} illustrate examples of HKAN's fitting for TF3 and TF5. Unlike TF2, these target functions were not affected by noise, enabling HKAN to achieve nearly perfect fitting with just two layers.

\begin{figure}[!t]
\centering
{\includegraphics[width=0.46\textwidth]{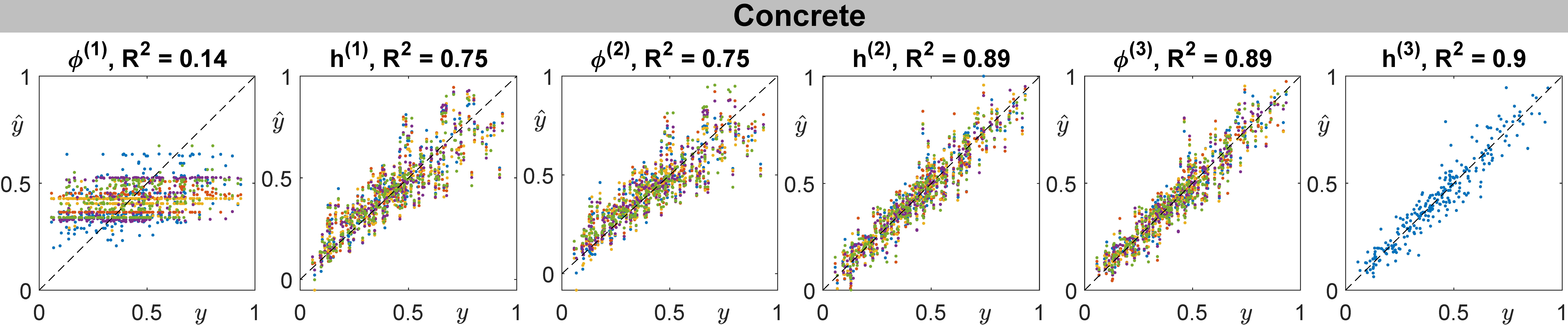}}
{\includegraphics[width=0.3020\textwidth]{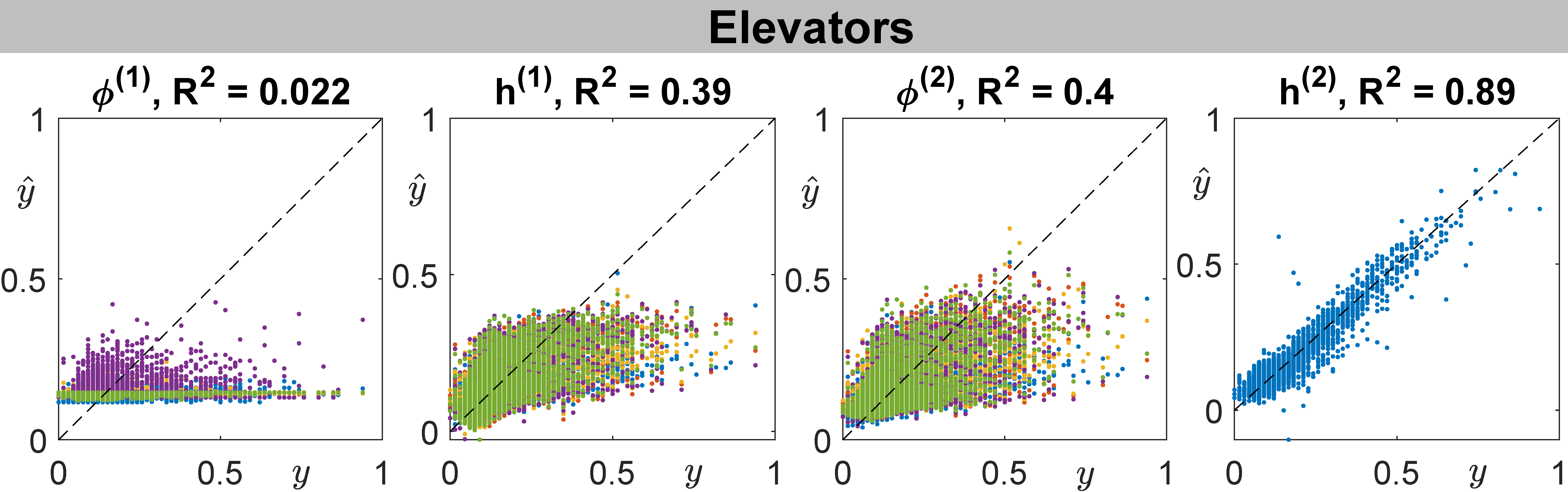}}
{\includegraphics[width=0.1540\textwidth]{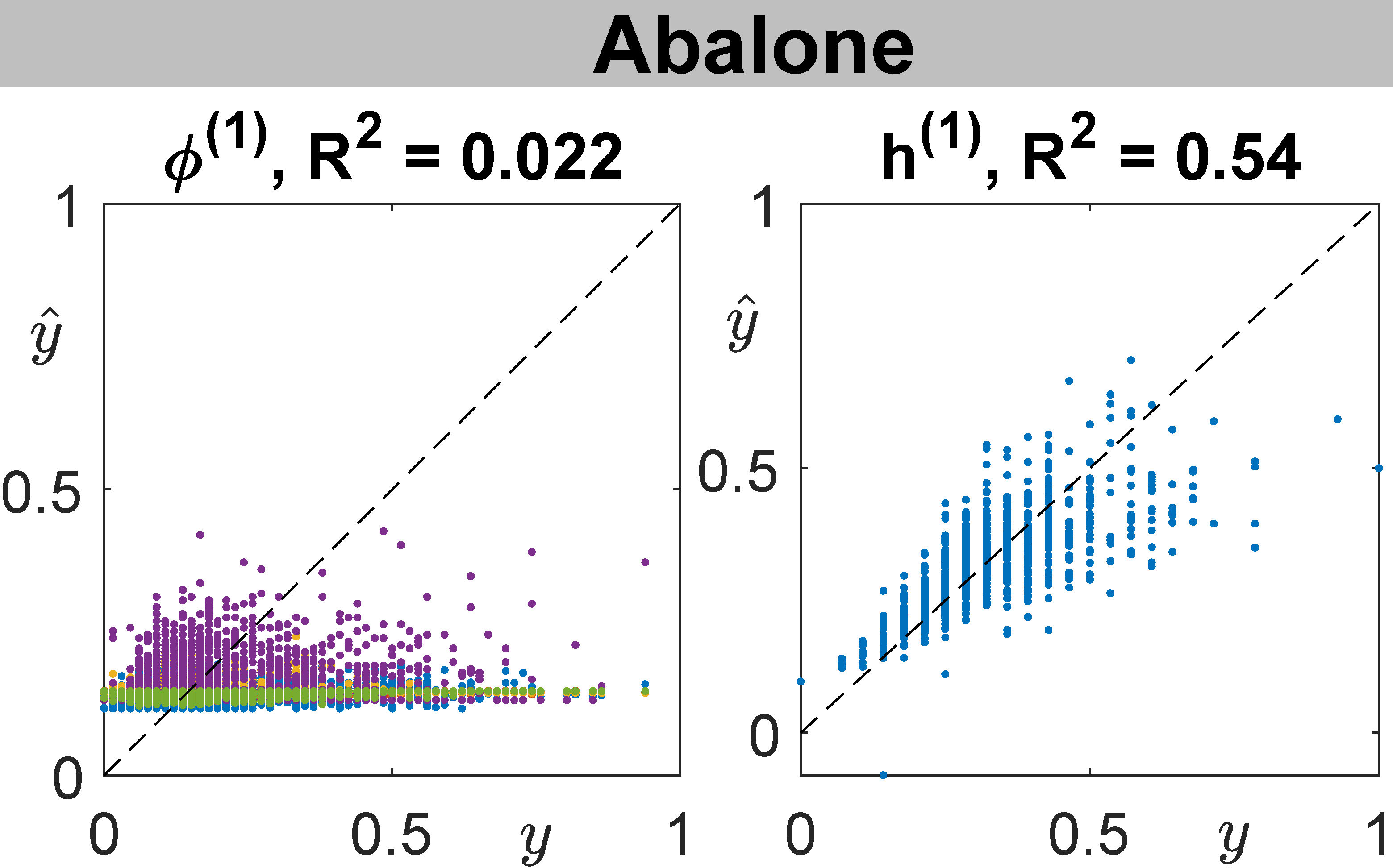}}
{\includegraphics[width=0.40\textwidth]{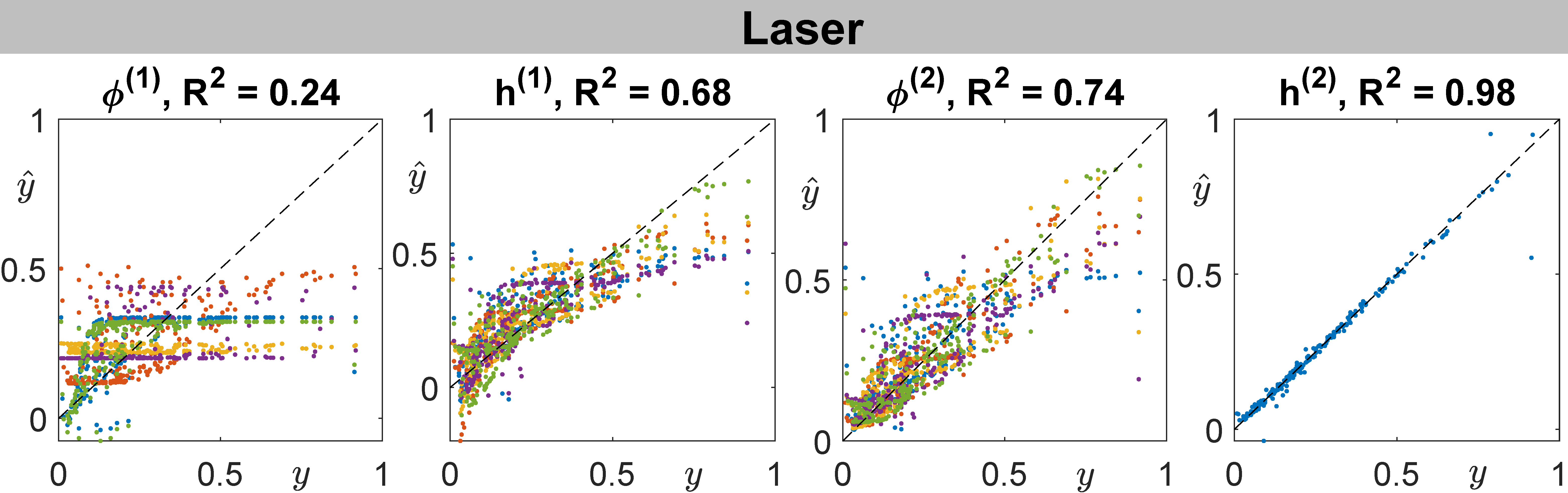}}
{\includegraphics[width=0.40\textwidth]{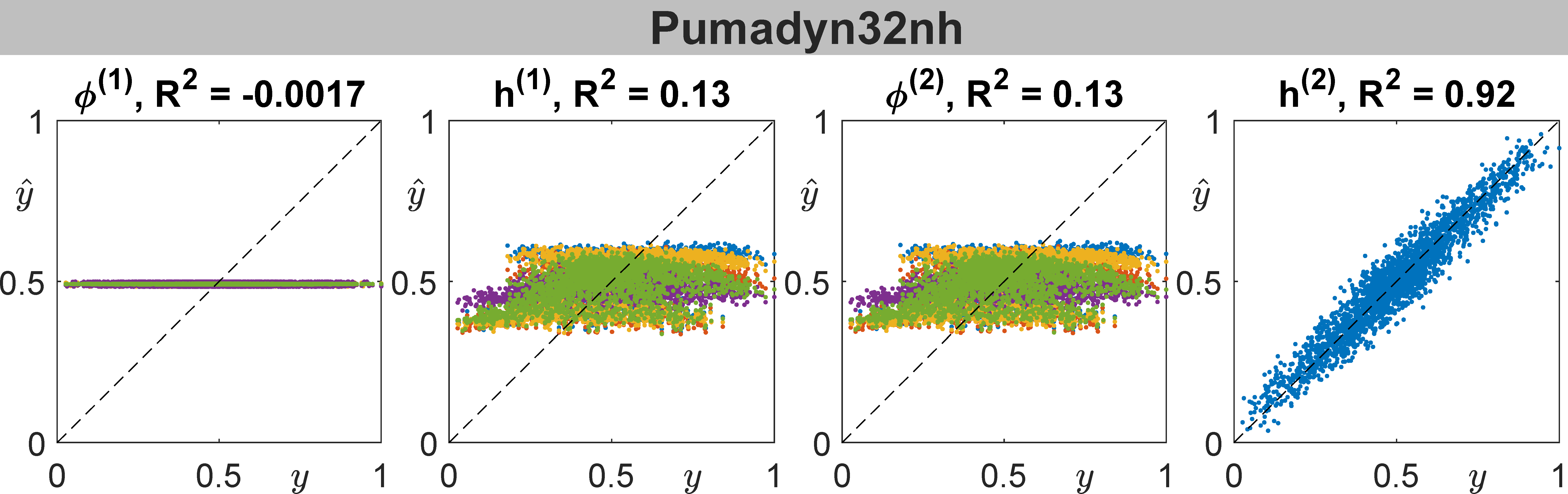}}
\caption{Predicted vs. target plots for selected datasets.}
\label{fig6d}
\end{figure}

Additional examples are presented in Fig. \ref{fig6d}. Among these, the Concrete dataset required the most complex architecture with three layers, while the Abalone dataset achieved its best fit with a simple architecture of just one layer. However, the latter case demonstrates that satisfactory results are not always guaranteed.

These visualizations emphasize HKAN's hierarchical modeling process, where prediction quality is progressively refined through successive layers, adapting to the complexity and structure of the target function.

\subsection{Input Argument Importance Estimation by HKAN}


HKAN includes a built-in mechanism for estimating the importance of input arguments. The blocks in the first layer approximate the target function based on individual inputs, and the accuracy of each block's fitting, measured by $R^2$, serves as a proxy for the importance of the corresponding input. 
However, it should be noted that this importance is estimated based on the coarse approximation performed by the first-layer blocks. The more refined approximations developed in subsequent layers are not considered in this estimation.

\begin{figure}[!t]
\centering
\includegraphics[width=0.48\textwidth]{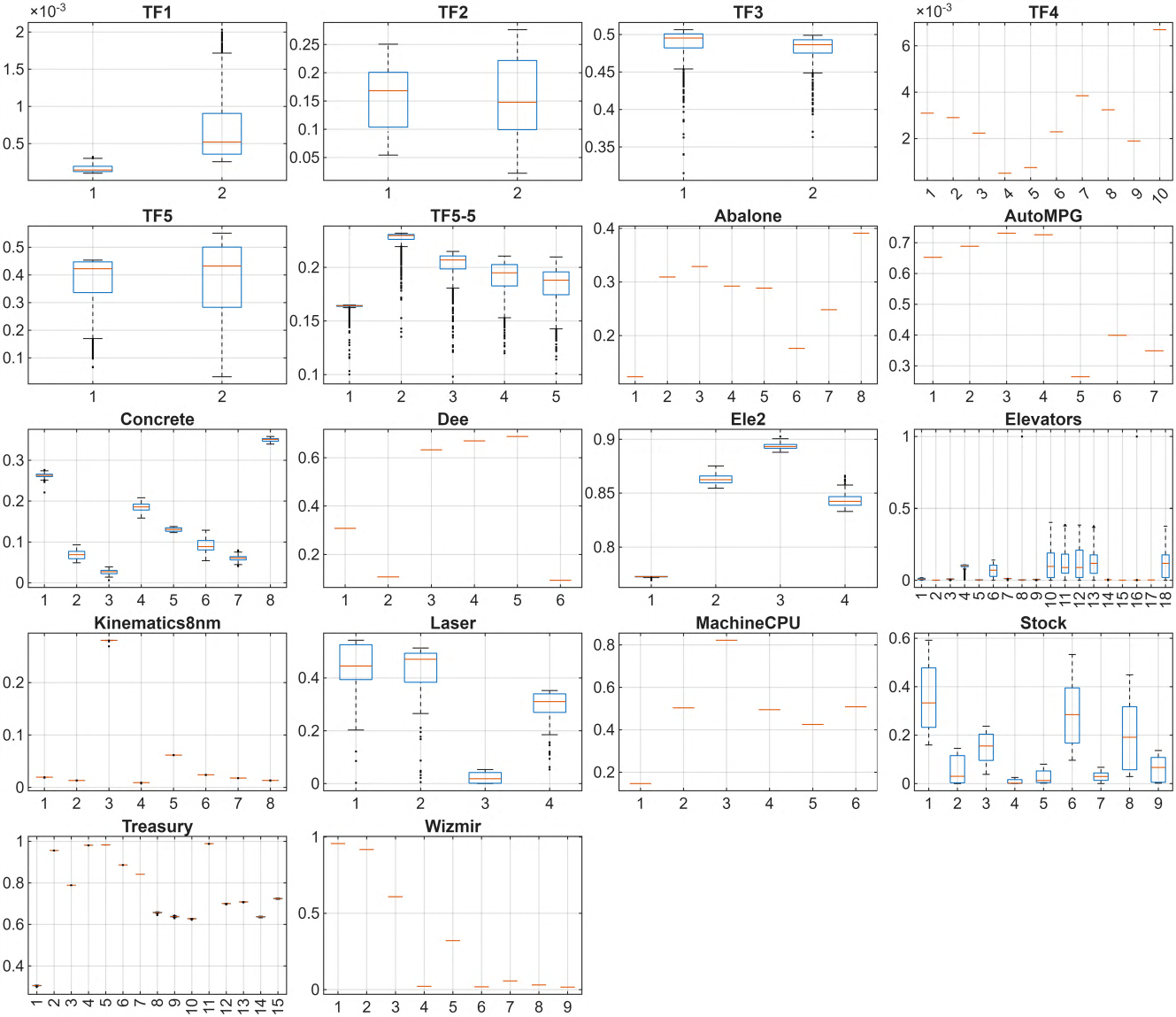}
\includegraphics[width=0.48\textwidth]{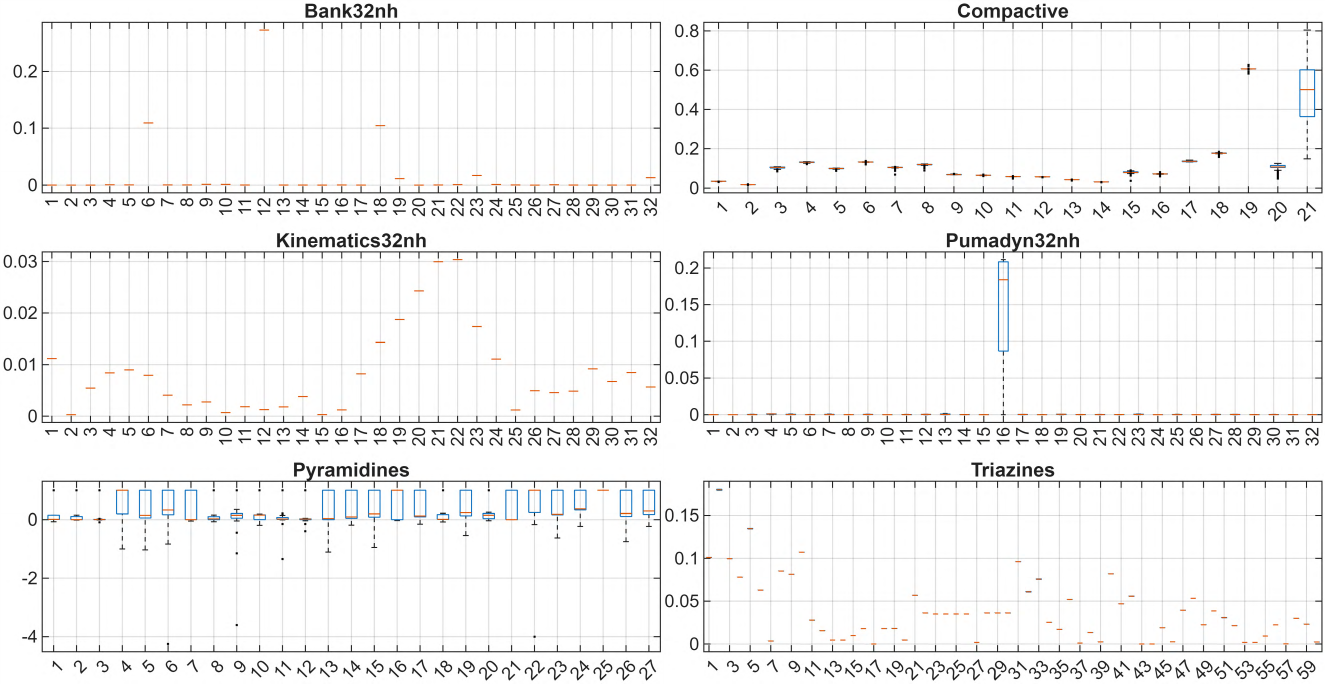}
\caption{Input arguments importance: $R^2$ for first-layer BlFs.}
\label{fig7}
\end{figure}

Fig. \ref{fig7} presents boxplots of the $R^2$ values for predictions made by the first-layer blocks. In some cases, such as TF1 and TF4, all 
$R^2$ values are very small (less than 0.01), indicating that the blocks provide a very weak approximation of the target function. By contrast, significantly higher $R^2$ values observed for other synthetic functions highlight a more balanced importance across input arguments, which aligns with expectations.

The greatest variation in $R^2$ values among input variables, exceeding 0.5, is observed for datasets such as Compactive, Dee, MachineCPU, Pyramidines, Stock, Treasury, and Wizmir. In these cases, large differences in variable importance are evident even during the coarse modeling of the target function by the first-layer blocks.

The average importance of the $p$-th input argument can be estimated using the BlFs associated with this argument as their average $R^2$:

\begin{equation}
\label{eqIp}
I_p = \frac{1}{n^{(1)}}\sum_{q=1}^{n^{(1)}} R^2(y,\phi_{q,p}^{(1)})
\end{equation}




\subsection{Discussion} \label{Disc}

{

The experimental results highlight HKAN's potential as a robust alternative to backpropagation-based KAN. Its hierarchical multi-stacking approach and randomized learning process make it particularly well-suited for applications that require rapid model deployment and transparency in variable importance.

\subsubsection{\textbf{Learning Process}}

HKAN eliminates the iterative backpropagation process, in many cases significantly reducing  training time compared to KAN and MLP (see Table \ref{tab4b}).
By transforming the optimization problem into multiple convex subproblems solved using least-squares regression, HKAN ensures efficient training while maintaining accuracy. Its deterministic training process enhances stability, while the layer-by-layer hierarchical approach facilitates transparent function representation. The flexibility of HKAN's architecture offers a notable advantage in capturing diverse functional relationships within the data.

\subsubsection{\textbf{Hierarchical Approximation of the Target Function}}

In HKAN, the fitted functions are constructed hierarchically, evolving in shape and complexity across layers. An initial layer focuses on features derived from individual input variables, while subsequent layers integrate these approximations across all variables and progressively refine their quality. As the network deepens, modeling variance decreases significantly, contributing to the accuracy and stability of the model's predictions.

\subsubsection{\textbf{Built-in Ensembling}}

HKAN incorporates two ensemble-inspired principles that enhance its generalization ability and robustness:

\begin{itemize}
\item Horizontal integration: Within each layer, diverse blocks are combined, similar to ensemble methods. Each block provides a unique perspective on the input data, and their aggregation enables the network to capture a broader spectrum of patterns and dependencies.
\item Vertical integration: The layer-by-layer processing corresponds to multi-level stacking, where each successive layer builds upon the approximations learned by the previous one. This hierarchical composition allows the network to construct increasingly complex functional representations.
\end{itemize}

Unlike KAN, which typically employs evenly distributed BaFs \cite{Liu24}, HKAN uses data-driven or randomized BaF distributions. This design promotes diversity among blocks and prevents them from converging to identical parameterizations. The blocks can thus be regarded as weak learners that are aggregated through stacking. In this context, inducing diversity among learners is deliberate and crucial for reinforcing the ensemble effect.

These ensemble-like mechanisms enable HKAN to capture intricate dependencies within the data without relying on backpropagation, offering a distinctive and computationally efficient framework for function approximation and pattern recognition tasks.

\subsubsection{\textbf{Flexibility}}

The complexity of the target function directly determines the optimal architecture of HKAN. More complex target functions require deeper and wider networks, whereas simpler functions can be effectively modeled with fewer layers, blocks, and BaFs.
A distinctive feature of HKAN is its flexibility in selecting BaFs. While KAN relies on B-splines \cite{Liu24}, which can be computationally demanding due to their recursive nature, HKAN supports alternative basis functions such as Gaussian and sigmoid, among others.
The smoothing parameter in HKAN serves as another critical hyperparameter, offering additional control over the model’s flexibility and adaptability.

\subsubsection{\textbf{Built-in Mechanism for Estimating Input Importance}}

A key strength of HKAN is its built-in mechanism for estimating the importance of input arguments, providing valuable insights into the significance of individual variables. This feature enhances interpretability, making HKAN particularly useful in applications where understanding variable contributions is crucial. In contrast, the interpretability of KAN stems from its ability to explicitly model functional relationships between input arguments and the output variable, enabling an understanding of both the nature and extent of each input's influence on predictions.

Compared with KAN, HKAN provides a more limited, feature-level interpretability, focusing on the analysis of univariate component functions and input-variable importance rather than symbolic interpretability.

\subsubsection{\textbf{Explainability}}

HKAN exhibits a high degree of explainability due to its transparent, hierarchical, and modular structure. Unlike end-to-end gradient-based NNs, HKAN decomposes the learning process into a sequence of convex subproblems, each corresponding to an explicit functional transformation. This structure allows each model component to be individually analyzed and interpreted.

Key factors contributing to HKAN’s explainability include:

\begin{itemize}
\item {Transparent hierarchical structure}: HKAN constructs the target function layer by layer, allowing detailed examination of each stage of approximation (see Figs. \ref{fig6a}-\ref{fig6d}).
\item {Univariate blocks and built-in input importance estimation}: Eeach block models the relationship between a single input and the output, making input effects easy to interpret. HKAN inherently quantifies the contribution of each input variable using metrics such as $R^2$.
\item {Deterministic learning process}: The absence of stochastic gradient updates ensures reproducible and stable results, enhancing interpretability (assuming fixed BaF parameters).
\item {Explicit functional form}: HKAN’s output is represented as nested linear combinations of known basis functions, allowing direct tracing of how each component contributes to the final prediction.
\end{itemize}

\subsubsection{\textbf{Relation to Adaptive and Ensemble-Based Models}}

HKAN can also be situated within the broader family of adaptive, aggregating, and expanding models. Classical approaches such as Boosting, Additive Trees, and Multivariate Adaptive Regression Splines (MARS) \cite{Has17} also avoid strict parametric assumptions and rely on flexible function approximation. What these models share with HKAN is the principle of constructing predictive functions from relatively simple components -- be they weak learners in Boosting, piecewise trees in Additive Models, or spline basis functions in MARS -- and combining them in a structured way to achieve high predictive accuracy.

Despite this high-level similarity, HKAN is fundamentally different from these methods both in architecture and in training procedure. Boosting and Additive Trees rely on sequentially improving ensembles of decision trees using stage-wise optimization, while MARS constructs adaptive spline bases through forward-backward selection. In contrast, HKAN employs randomized basis functions and performs training exclusively through closed-form least-squares regression without iterative optimization. This results in a deterministic, convex learning procedure that clearly distinguishes HKAN from the aforementioned methods.

By combining principles of randomized learning, hierarchical stacking, and functional approximation rooted in the Kolmogorov-Arnold framework, HKAN constitutes a novel contribution that goes beyond existing approaches.

}








\section{Conclusion}

This study introduces the Hierarchical Kolmogorov-Arnold Network as an efficient and interpretable alternative to traditional backpropagation-based NNs, particularly KAN. By employing a randomized learning approach based on linear regression and a hierarchical multi-stacking architecture, HKAN eliminates the need for iterative gradient-based training, reducing {training time} while maintaining or even enhancing accuracy and stability.  

The empirical evaluation demonstrates that HKAN performs competitively across diverse regression tasks, effectively capturing complex relationships within data. Additionally, its built-in mechanism for estimating input variable importance enhances interpretability, making it a valuable tool for applications requiring transparency and explainability.  
The flexibility of HKAN in terms of basis functions and architecture allows it to adapt to varying complexities of target functions, further establishing its potential for real-world applications. 

Future research could explore extending HKAN's capabilities to classification and forecasting tasks, as well as investigating its integration with other advanced architectures to expand its applicability.  

{
\section*{Acknowledgments}
The authors would like to thank the anonymous reviewer for the valuable comments on the early developments of KANs. These insights helped us recognize that the origins of KANs trace back much earlier than the work presented in \cite{Liu24}.
}

{
\section*{Declaration of generative AI and AI-assisted technologies in the writing process}

During the preparation of this work the authors used AI-based language tools (ChatGPT, Claude, and Gemini) in order to improve language and readability. After using this tools/services, the authors reviewed and edited the content as needed and take full responsibility for the content of the publication.
}


\end{document}